\documentclass[journal,twoside,web]{ieeecolor}
\usepackage{generic}
\usepackage{cite}
\usepackage{amsmath,amssymb,amsfonts}
\usepackage{algorithm}

\usepackage{algpseudocode}
\usepackage{amsmath}
\usepackage{graphicx}
\usepackage{multirow}
\usepackage{textcomp}
\usepackage{array}
\usepackage{makecell}
\usepackage{CJKutf8}
\usepackage{color}
\usepackage{soul}
\usepackage[export]{adjustbox} 
\usepackage{bm}
\usepackage{tabularx}
\usepackage{caption}
\usepackage{booktabs} 
\usepackage{siunitx}
\usepackage{subcaption}
\soulregister\cite7 
\soulregister\citep7 
\soulregister\citet7 
\soulregister\ref7 
\soulregister\pageref7 
\makeatletter
\let\NAT@parse\undefined
\makeatother
\usepackage[colorlinks=true,
           linkcolor=blue,
           anchorcolor=blue,
           citecolor=green,
           urlcolor=blue,
           ]{hyperref}
\usepackage[numbers,sort&compress]{natbib}

\def\BibTeX{{\rm B\kern-.05em{\sc i\kern-.025em b}\kern-.08em
    T\kern-.1667em\lower.7ex\hbox{E}\kern-.125emX}}
\begin{document}

\title{Towards Accurate and Lightweight Peripheral Neuroblastic Tumor Diagnosis via Contrastive Multi-scale Pathological Image Analysis}
\author{Zhu Zhu, Shuo Jiang, Jingyuan Zheng, Yawen Li, Yifei Chen,\\ Manli Zhao, Weizhong Gu, Feiwei Qin, Jinhu Wang and Gang Yu
\thanks{This work was supported by the National Key Research and Development Program of China under Grant 2023YFC2706400, the National Natural Science Foundation of China under Grant No. 62576314, and the Fundamental Research Funds for the Provincial Universities of Zhejiang under Grant GK259909299001-006. (Zhu Zhu, Shuo Jiang, and Jingyuan Zheng contributed equally to this work.) (Corresponding authors: Feiwei Qin; Jinhu Wang; Gang Yu.)}
\thanks{Zhu Zhu, Manli Zhao, Weizhong Gu, Jinhu Wang and Gang Yu are with Children's Hospital, Zhejiang University School of Medicine, Hangzhou, 310000, China (e-mail: \{zhuzhu$\_$cs; zhaomanli; 6195013; wjh; yugbme\}@zju.edu.cn).}
\thanks{Shuo Jiang, Jingyuan Zheng and Feiwei Qin are with the Hangzhou Dianzi University, Hangzhou 310018, China (e-mail: \{jiangshuo; zhengjoy; qinfeiwei\}@hdu.edu.cn).}
\thanks{Yawen Li is with the Chinese University of Hong Kong, Hong Kong SAR, China (e-mail: 1155246089@link.cuhk.edu.hk).}
\thanks{Yifei Chen is with Tsinghua University, Beijing 100084, China (e-mail: justlfc03@gmail.com).}
}


\maketitle

\begin{abstract}
Peripheral neuroblastic tumors (pNTs) are among the most common extracranial solid tumors in children, and accurate pathological subtyping is important for risk stratification and treatment planning. However, pNT subtyping on hematoxylin-eosin whole-slide images (WSIs) remains challenging because of limited pediatric tumor cohorts, marked histological heterogeneity, inter-observer variability, and the computational burden of existing WSI classifiers. To address these challenges, we propose CoPath, a framework consisting of CoHisNet and PathVote. CoHisNet is a lightweight multi-scale feature-fusion network for patch-level histopathological classification. By replacing the multilayer perceptron components in Swin Transformer blocks and the classification head with Kolmogorov-Arnold Network layers, CoHisNet improves nonlinear feature modeling under a compact architecture. Its multi-scale interaction and contrast-driven feature-enhancement design enables the model to capture both tissue-level structures and fine-grained cellular morphology. PathVote further incorporates pathology-informed tissue-component priors to aggregate patch-level predictions into WSI-level decisions. We validated CoPath on a private two-branch PpNTs cohort and the public BreakHis breast cancer histopathology dataset. Experimental results show that CoPath achieves competitive or superior performance compared with general image classifiers, pathology foundation models under linear probing, and pathology-specific classification models, while maintaining substantially lower computational complexity. The source code is available at \href{https://github.com/JSLiam94/CoPath}{https://github.com/JSLiam94/CoPath}.
\end{abstract}


\section{Introduction}
\label{sec:introduction}
\IEEEPARstart{P}{eripheral} neuroblastic tumors (pNTs), originating from the developing sympathetic nervous system, are among the most common extracranial solid tumors in children \cite{bagatell2024neuroblastoma}. They account for approximately 15\% of cancer-related deaths in children \cite{fetahu2023single}. pNTs comprise pathological subtypes, each showing disparities in tissue morphology, malignancy, and treatment strategies. Thus, achieving accurate subtype classification is essential for risk stratification and individualized treatment. Currently, histopathological examination of H\&E-stained tumor tissue under light microscopy remains the diagnostic gold standard for pNTs. In pathological assessment, pNTs are classified according to the International Neuroblastoma Pathology Classification (INPC), which is based on the Shimada system and has been internationally validated for prognostic relevance~\cite{shimada1999terminology,shimada2012international}. The diagnostic process requires the evaluation of neuropil, Schwannian stromal development, nested cellular architecture, and the proportion of differentiated neuroblasts. However, these morphological criteria are assessed by pathologists through examination of WSIs and therefore may be affected by inter-observer variability, especially in cases with heterogeneous tumor composition. Therefore, computational tools capable of identifying diagnostically relevant morphology and supporting reproducible pNT subtyping are of clinical significance.

In recent years, substantial progress has been achieved in the analysis of pathological images using deep learning. Approaches such as multi-instance learning (MIL) and visual Transformers have been extensively applied to cancer classification, staging, and prognosis analysis based on WSIs \cite{xu2023vision, liu2024pseudo, wang2024deep, yan2025multi}. Nevertheless, directly applying these methods to the fine-grained subtyping of pNTs still poses numerous challenges: First, current supervised learning approaches rely significantly on a substantial amount of pixel-level or region-level labeled data \cite{coudray2018classification}. However, acquiring fine-grained annotations for pNTs is extremely expensive, and the sample size is restricted. This severely limits the training and application of fully supervised models. Weakly supervised methods, such as CLAM \cite{lu2021data} and ClassicMIL \cite{campanella2019clinical}, although reducing the dependence on annotations, still necessitate thousands of WSIs to achieve optimal performance. This creates an insurmountable data scale bottleneck for rare diseases such as pNTs. Second, pNTs exhibit substantial histological heterogeneity. According to the INPC-based assessment, diagnostically relevant components include neuropil, neuroblastic tumor cells with different degrees of differentiation, mature ganglion cells, nested neuroblastic structures, and Schwannian stroma. These components may coexist within the same WSI and show variable spatial distribution. Therefore, an effective model should integrate local cellular morphology with broader tissue context, rather than relying only on independent patch-level predictions. Traditional MIL methods treat image patches as independent instances~\cite{lu2021ai} and lack the capacity to model the global tissue structure context, making it difficult to capture long-range dependencies and cross-region interactions \cite{yu2024patch, NEURIPS2021Shao}, which can lead to a decline in classification performance. Third, although Transformer models show advantages in capturing long-range dependencies and hierarchical context \cite{zheng2023kernel}, their high computational complexity and the general absence of lightweight design impede their deployment and application in clinical scenarios \cite{chen2024sckansformer, wani2024explainable}. Additionally, existing methods often lack interpretable decision-making mechanisms, making it difficult to incorporate prior pathological knowledge, which leads to insufficient trust in model results by clinicians \cite{10.1007/978-3-031-43987-2_10, 9380482}.


To address these challenges, we propose CoPath, a framework for fine-grained pNT subtyping on hematoxylin and eosin (H\&E)-stained WSIs. CoPath comprises the patch-level classifier CoHisNet and the WSI-level aggregation module PathVote. CoHisNet learns compact cross-scale representations that integrate fine-grained cellular morphology with broader tissue context, while PathVote incorporates pathology-informed tissue-component priors to aggregate patch-level predictions into slide-level decisions. The main contributions are summarized as follows:

\begin{itemize}
  \item \textbf{Contrast-driven multi-scale feature interaction:}
  We develop a Contrast-driven Multi-scale feature Interaction and Fusion (CMIF) module that combines adaptive scale weighting with contrast-driven feature enhancement. By explicitly interacting hierarchical features, CMIF balances information across scales and strengthens the representation of subtle morphological differences in histopathological images.

  \item \textbf{Compact KAN-enhanced architecture:}
  We introduce a Swin KANsformer backbone that retains regular and shifted-window self-attention while replacing conventional MLP sublayers with Kolmogorov--Arnold Network (KAN) sublayers. Together with a KAN-based classification head and reduced embedding dimensions, this design provides flexible nonlinear modeling under a compact computational configuration.

  \item \textbf{Pathology-informed WSI-level aggregation:}
We propose PathVote, an adaptive voting strategy that jointly incorporates patch confidence, prediction margins, and INPC-motivated tissue-component priors. A dedicated Prior Synergy Block estimates neuropil, Schwannian stroma, and other tissue components, enabling diagnostically informative and reliable patches to contribute differentially to the final WSI-level prediction.

\end{itemize}

\begin{figure*}[t]
    \centering
    \includegraphics[width=\textwidth]{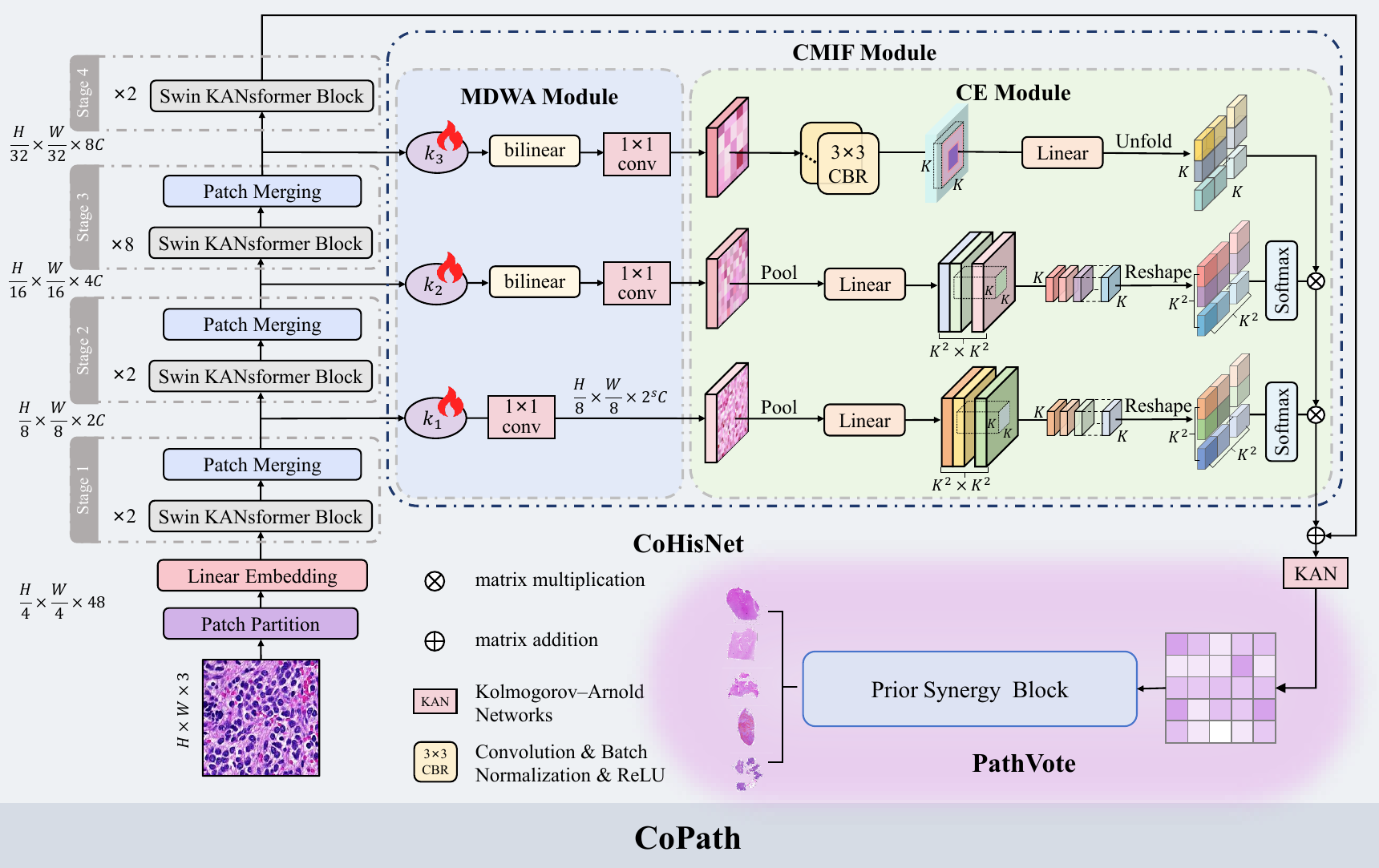}
\caption{Overview of CoPath. CoHisNet extracts hierarchical patch representations using Swin KANsformer blocks and fuses multi-scale features through CMIF. PathVote then combines patch-level logits, confidence margins, and tissue-component priors estimated by the Prior Synergy Block to produce the WSI-level prediction.}
    \label{fig:CoPath}
\end{figure*}

\section{Related work}
\subsection{Traditional Computer Vision Models in Pathology}

Early computer-aided pathology systems combined handcrafted descriptors with conventional classifiers. For neuroblastoma subtyping, Gheisari et al.~\cite{gheisari2018computer} combined Scale-Invariant Feature Transform descriptors with feature encoding and a support vector machine. Their Convolutional Deep Belief Network learned higher-level representations under the Shimada classification system \cite{gheisari2018convolutional}. Zhu et al.~\cite{zhu2026unified} introduced NEVA, a WSI-report vision-language model for neuroblastoma diagnosis and risk stratification. Although these studies established the feasibility of automated neuroblastoma assessment, handcrafted or shallow representations remain sensitive to staining and morphological variation and do not capture long-range tissue organization in high-resolution WSIs.

Patch-based convolutional neural networks (CNNs) subsequently became a practical means of processing gigapixel WSIs. Hou et al.~\cite{hou2016patch} aggregated local patch predictions through decision fusion, improving computational tractability but retaining an independent-patch representation that can fragment spatial context. Weakly supervised multiple-instance learning partly addresses this limitation: TransMIL models correlations among instances using a Transformer architecture \cite{NEURIPS2021Shao}, while variational-information-bottleneck fine-tuning suppresses task-irrelevant information during WSI representation learning \cite{li2023task}. Nevertheless, reliable global modeling remains difficult when diagnostic structures are spatially heterogeneous and only limited rare-disease cohorts are available.


\subsection{Development of Pathology Foundation Models}
Self-supervised learning has enabled pathology foundation models to learn transferable representations from large-scale histopathological data. CTransPath combines convolutional inductive bias with a multi-scale Swin Transformer and contrastive pretraining to capture both local texture patterns and hierarchical tissue context \cite{wang2022transformer}. UNI further demonstrates that large-scale, task-agnostic pretraining can support diverse downstream applications, including tissue recognition and cancer subtype classification \cite{chen2024towards}. Although these models provide broadly transferable features, their performance on specialized pediatric tumors remains dependent on the downstream adaptation strategy, the scale and composition of the target cohort, and the morphological correspondence between the pretraining data and the target disease.

Multimodal pretraining extends this paradigm. CONCH aligns histopathological images with natural-language supervision, enabling classification, segmentation, and retrieval within a shared representation space \cite{lu2024visual}. REMEDIS emphasizes self-supervised transfer across medical imaging tasks \cite{azizi2023robust}, whereas MUSK integrates visual and textual pathology information for precision-oncology applications \cite{Xiang2025MUSK}. Despite their generalization capacity, foundation models require substantial computational resources and may preserve semantic information that is not optimally aligned with subtle, fine-grained distinctions among pNT subtypes. This motivates fair comparison with compact task-specific models under identical downstream splits and probing protocols.



\subsection{Advantages of Lightweight Histopathology Models}

Lightweight models offer lower training and inference costs, easier deployment, and greater flexibility for disease-specific optimization. SuperHistopath uses superpixel segmentation and CNN classification to map tumor heterogeneity in low-resolution WSIs \cite{zormpas2021superhistopath}. StoHisNet combines CNNs and Transformers to integrate local appearance with global context \cite{fu2022stohisnet}, while HisImage incorporates wavelet-based positional encoding and external attention to improve feature continuity efficiently \cite{ding2023enhanced}. FMDNN introduces fuzzy-guided multi-granularity reasoning to approximate the hierarchical assessment performed by pathologists \cite{ding2024fmdnn}.

Other efficient designs focus on selecting or enhancing diagnostically informative features. ScoreNet learns non-uniform attention over pathological regions and couples it with task-oriented augmentation \cite{stegmuller2023scorenet}; DetexNet strengthens low-level texture representation for challenging pediatric tumors \cite{liu2020detexnet}. These approaches reduce dependence on very large backbones, yet simultaneously preserving nuclear detail, tissue architecture, and cross-scale context remains difficult. Moreover, parameter count alone does not establish practical efficiency, because throughput and memory use depend on the computational characteristics of individual operators.

\subsection{Multi-scale Methods in Medical Image Processing}
Multi-scale representation is central to medical image analysis because cellular morphology and tissue architecture provide complementary evidence across multiple spatial resolutions and contexts \cite{DU2020105603}. CrossViT exchanges information between branches operating on patches of different sizes through cross-attention \cite{chen2021crossvit}. MViTv2 constructs hierarchical features using relative position encoding and residual pooling \cite{li2022mvitv2}, Shunted Self-Attention aggregates tokens through heterogeneous receptive fields \cite{ren2022shunted}, and DilateFormer expands contextual coverage using multi-scale dilated attention \cite{jiao2023dilateformer}. Collectively, these methods show that explicit communication across scales is preferable to independently processing features and concatenating them only at the output.


In medical imaging, HiFuse adaptively combines global and local branches for classification \cite{huo2023hifuse}, and multilevel deformable-DETR features improve the characterization of leukocytes at different spatial resolutions \cite{chen2024accurate}. ConDSeg further introduces contrast-driven feature enhancement to emphasize informative differences across representations and improve boundary recognition \cite{lei2024condseg}. However, existing multi-scale methods are not specifically designed to combine fine-grained pNT morphology, compact nonlinear modeling, and pathology-informed WSI aggregation.


\section{Method}
\label{sec:guidelines}

\subsection{Overall Architecture}
Fig.~\ref{fig:CoPath} illustrates the overall architecture of CoPath, which comprises the patch-level backbone CoHisNet and the WSI-level aggregation module PathVote. CoHisNet employs Swin KANsformer blocks and the Contrast-driven Multi-scale feature Interaction and Fusion (CMIF) module to learn compact cross-scale representations from H\&E-stained pathological patches, integrating fine-grained cellular morphology with broader tissue context. PathVote subsequently converts the patch-level outputs into WSI-level predictions by combining predictive confidence, confidence margins, and INPC-motivated tissue-component priors. Unlike many conventional two-stage MIL pipelines that aggregate pre-extracted patch embeddings, CoPath combines task-specific patch representation learning with pathology-informed slide-level aggregation, thereby addressing both fine-grained morphological modeling and interpretable WSI-level decision-making.

\subsection{Swin KANsformer Block}

\begin{figure}[htbp]
    \centering
    \includegraphics[width=\columnwidth]{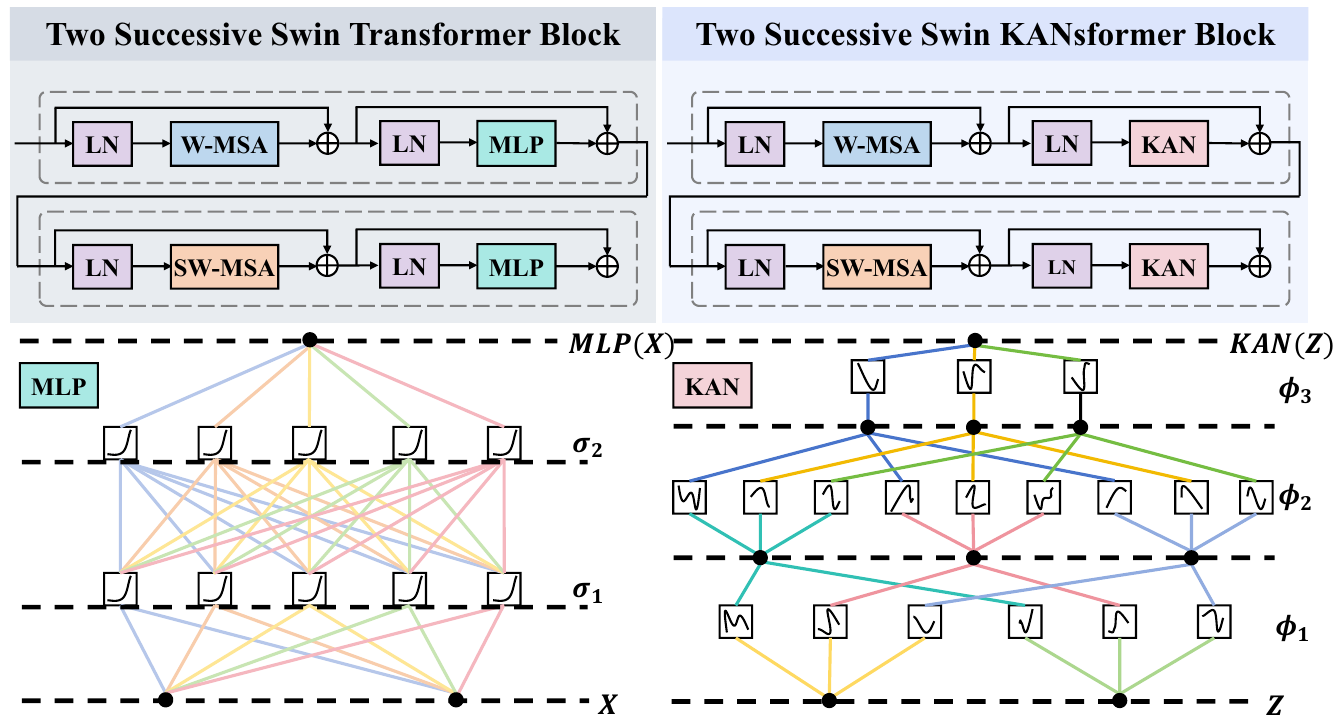}
    \caption{Structural comparison between the Swin Transformer block and the proposed Swin KANsformer block. The Swin KANsformer retains the window-based attention architecture while replacing each conventional MLP sublayer with a KAN sublayer parameterized by learnable spline functions.}
    \label{fig:kans}
\end{figure}

The Swin Transformer constructs hierarchical representations through local window-based self-attention, serving as a backbone for high-resolution histopathological image analysis. As shown in Fig.~\ref{fig:kans}, we develop the Swin KANsformer block by replacing each conventional multilayer perceptron (MLP) sublayer with a Kolmogorov-Arnold Network (KAN) sublayer \cite{liu2025kan}, while retaining layer normalization, residual connections, and the shifted-window attention mechanism. Consecutive blocks alternate between Window Multi-Head Self-Attention (W-MSA) and Shifted Window Multi-Head Self-Attention (SW-MSA), enabling local feature modeling within individual windows and information exchange across adjacent windows.

Given an input H\&E-stained image of size $H \times W \times 3$, the Patch Partition module divides it into non-overlapping $4 \times 4$ patches, producing $\frac{H}{4} \times \frac{W}{4}$ patch tokens, each with a feature dimension of $4 \times 4 \times 3=48$. A linear embedding layer subsequently projects each token into a $C$-dimensional representation. The backbone comprises four hierarchical stages, with the number of Swin KANsformer blocks defined as
$[d_0,d_1,d_2,d_3]=[2,2,8,2]$.
Each of the first three stages contains $d_i$ Swin KANsformer blocks followed by a patch-merging layer, whereas the final stage contains only $d_3$ blocks. Each patch-merging operation halves the spatial resolution and doubles the channel dimension. Consequently, the hierarchical feature maps generated after the first three stages have dimensions
$\left[
\frac{H}{8}\times\frac{W}{8}\times2C,\,
\frac{H}{16}\times\frac{W}{16}\times4C,\,
\frac{H}{32}\times\frac{W}{32}\times8C
\right].$
These feature maps capture complementary morphological information at progressively enlarged receptive fields and are retained as inputs to the CMIF module. The fourth stage further processes the deepest representation for subsequent residual fusion and patch-level classification.

\subsection{Contrast-Driven Multi-Scale Feature Interaction and Fusion Module}

The CMIF module integrates hierarchical representations through Multi-scale Dynamic Weight Adjustment (MDWA) and contrast-driven feature enhancement (CE). MDWA aligns and reweights features across stages, whereas CE uses shallow-scale attention to refine deeper representations.

\textbf{MDWA module.}
Let $\mathbf{F}_1$, $\mathbf{F}_2$, and $\mathbf{F}_3$ denote the feature maps retained from the first three Swin KANsformer stages, with spatial resolutions $\left(\frac{H}{8},\frac{W}{8}\right)$, $\left(\frac{H}{16},\frac{W}{16}\right)$, and $\left(\frac{H}{32},\frac{W}{32}\right)$ and channel dimensions $2C$, $4C$, and $8C$, respectively. MDWA assigns learnable scale weights $\left[k_1,k_2,k_3\right]$ to adaptively modulate the contribution of each stage. The second- and third-stage features are upsampled by bilinear interpolation to the spatial resolution of $\mathbf{F}_1$, while separate $1\times1$ convolutional layers project all three feature maps to a unified channel dimension of $2^{s}C$. In our implementation, $s=1$, yielding aligned feature maps of size $\frac{H}{8}\times\frac{W}{8}\times2C$ before passing to the CE module. This configuration preserves the standard hierarchical channel progression of the Swin KANsformer backbone while providing a compact representation for cross-scale fusion.

\textbf{CE module.} In the CE module, at each spatial location $(i,j)$, attention weights within a $K\times K$ window centered at $(i,j)$ are computed by combining details from shallow and deep features. Specifically, the processed shallow features (features retained from the 1st and 2nd stages) $F_1$ and $F_2$ are passed through two different linear layers to compute the attention weights $A_1$ and $A_2$, where $A_1,A_2\in\mathbb{R}^{\frac{H}{8}\times\frac{W}{8}\times K^2}$. The attention weight computation can be expressed as:
\begin{equation}
A_1=W_1\cdot F_1,A_2=W_2\cdot F_2,
\end{equation}
where $W_1,\ W_2\in\mathbb{R}^{2C\times K^4}$ are the linear transformation weight layers for the multi-scale feature maps from the 1st and 2nd stages, respectively. Each spatial location $(i,j)$ in the two attention maps is then reshaped into ${\hat{A}}_{1_{i,j}}\in\mathbb{R}^{K\times K\times K^2}$ and ${\hat{A}}_{2_{i,j}}\in\mathbb{R}^{K\times K\times K^2}$, both activated by softmax.

The processed deep feature (retained from the 3rd stage) $F_3\in\mathbb{R}^{\frac{H}{8}\times\frac{W}{8}\times2C}$ is initially fused through two CBR blocks. A CBR block combines a convolution layer, batch normalization, and a ReLU activation function, simplifying the network construction process and improving training efficiency and performance. The fused $2C$-dimensional feature vector is then projected through a linear layer to a value vector $\bm{V}_3\in\mathbb{R}^{\frac{H}{8}\times\frac{W}{8}\times2C}$. This value vector $\bm{V}_3$ is then unfolded over local windows, aggregating neighborhood information for each position. Let $\bm{V}_{\Delta_{i,j}}\in{R^{2C\times{K^2}}}$ represent all values within a local window centered at $\left(i,j\right)$, defined as:
\begin{equation}
\bm{V}_{\Delta_{i,j}}=\{\bm{V}_{i+p-\left\lfloor\frac{K}{2}\right\rfloor,j+q-\left\lfloor\frac{K}{2}\right\rfloor}\},\qquad0\leq p,q<K.
\end{equation}
The aggregation is performed in two steps:
\begin{equation}
\widehat{V}_{\Delta_{i,j}}
=
\operatorname{Softmax}\!\left(\widehat{A}_{1i,j}\right)
\otimes
\left[
\operatorname{Softmax}\!\left(\widehat{A}_{2i,j}\right)
\otimes
V_{\Delta_{i,j}}
\right].
\end{equation}
where $\otimes$ denotes matrix multiplication for local attention weighting. Finally, the weighted feature values are densely aggregated to produce the final output feature map. Specifically, the aggregated feature at position $(i,j)$ is:
\begin{equation}
\widetilde{F}_{i,j}
=
\sum_{m=0}^{K-1}
\sum_{n=0}^{K-1}
\widetilde{V}^{\,i,j}_{
\Delta_{\,i+m-\lfloor K/2\rfloor,\,
j+n-\lfloor K/2\rfloor}
}.
\end{equation}

\subsection{KAN Classification Head}

The multi-scale representation produced by the CMIF module is fused with the output of the fourth Swin KANsformer stage and subsequently passed to a KAN-based classification head. In contrast to conventional linear or MLP-based classifiers, KAN parameterizes each univariate transformation using a fixed basis function and a learnable spline function~\cite{liu2025kan}:
\begin{equation}
\phi(x)
=
w_b b(x)
+
w_s \sum_{i=0}^{n-1} c_i N_{i,k}(x),
\end{equation}
where $b(x)$ is a fixed basis function, such as SiLU; $w_b$ and $w_s$ are learnable scaling coefficients; $c_i$ denotes the learnable spline coefficient; and $N_{i,k}(x)$ is the $i$-th B-spline basis function of degree $k$.

Given a knot sequence $t=\{t_0,t_1,\ldots,t_n\}$, the B-spline basis functions are recursively defined for numerical evaluation using the Cox-de Boor formulation:
\begin{equation}
N_{i,0}(x)
=
\begin{cases}
1, & t_i \leq x < t_{i+1},\\
0, & \text{otherwise},
\end{cases}
\end{equation}
and, for $k\geq1$,
\begin{equation}
\begin{aligned}
N_{i,k}(x)
={}&
\frac{x-t_i}{t_{i+k}-t_i}N_{i,k-1}(x)\\
&+
\frac{t_{i+k+1}-x}{t_{i+k+1}-t_{i+1}}
N_{i+1,k-1}(x).
\end{aligned}
\end{equation}
The local support of B-spline bases confines each response to a limited interval of the knot sequence, thereby providing a flexible and efficient piecewise-polynomial parameterization for patch-level classification.

\begin{figure}[htbp]
    \centering
    \includegraphics[width=\columnwidth]{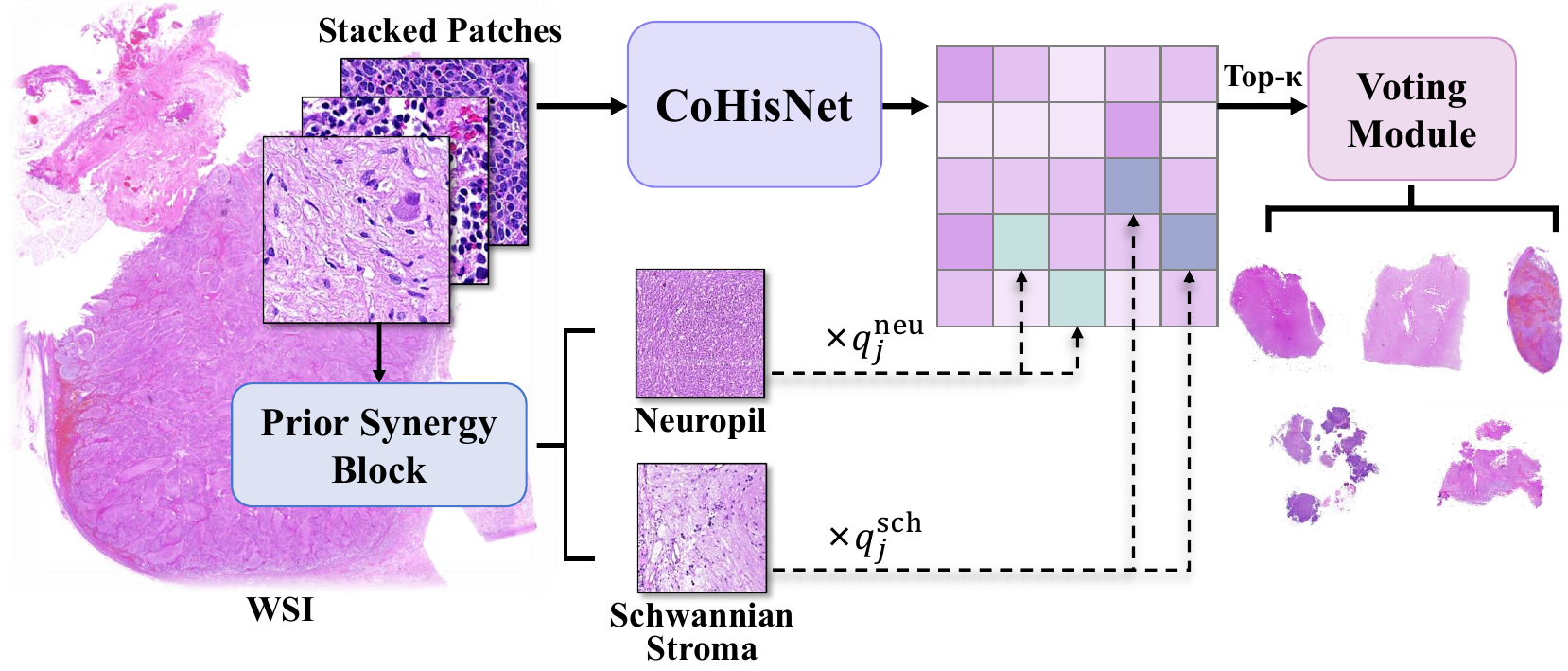}
    \caption{Architecture of PathVote. The original H\&E WSI is standardized,
divided, and stacked into patches. The Prior Synergy Block predicts the
tissue component of each selected patch, and category-specific prior
weights are incorporated into the subsequent weighted soft-voting
procedure to obtain the final WSI-level prediction.}
    \label{fig:wsi-vote}
    \vspace{-1.0em}
\end{figure}

\subsection{Domain Knowledge Guided Heuristic WSI Fine-Grained Classification Mechanism}
For WSI-level pathological diagnosis, we propose PathVote, a heuristic soft-voting classification pipeline that combines individual patch classifications with global feature information, using weighted patch importance to yield a final label that is more robust than hard voting, as illustrated in Fig. \ref{fig:wsi-vote}.

\subsubsection{Medical Knowledge Guided Heuristic Voting Mechanism Design}

WSI-level hard voting assigns equal importance to all selected patches, although pNT subtyping depends on specific histomorphological components rather than patch frequency alone across the whole slide, especially in heterogeneous regions. Under the INPC framework, neuropil provides evidence of neuroblastic differentiation, whereas Schwannian stromal development is important for distinguishing stroma-poor from stroma-rich or stroma-dominant tumors. Notably, the 50\% criterion in INPC refers to the tissue-level proportion of Schwannian stroma and tumor components, rather than the area proportion within an individual patch. PathVote therefore does not impose a hard 50\% threshold on patch predictions; instead, neuropil and Schwannian stroma are incorporated as probabilistic tissue-component priors during WSI-level aggregation.
Selected patches are assigned confidence-, margin-, and tissue-prior-dependent weights before slide-level aggregation.


\subsubsection{Training and Application of the Prior Synergy Block}

Leveraging the features extracted by CoHisNet, the Prior Synergy Block
is trained to predict whether a patch contains neuropil, Schwannian
stroma, or other tissue components. For each selected patch $j$, let
$\mathbf{z}_j \in \mathbb{R}^{C}$ denote the class-logit vector predicted
by CoHisNet, where $C$ is the number of diagnostic categories, and let
$
\mathbf{p}_j=\mathrm{softmax}(\mathbf{z}_j)
$
denote the corresponding class-probability vector. The patch confidence
and predicted patch labels are defined as

\begin{equation}
\mathrm{conf}_j=\max_{c}p_{j,c},
\qquad
\hat{y}_j=\arg\max_{c}p_{j,c}.
\label{eq:patch_confidence}
\end{equation}

To quantify the discriminative reliability of each patch, we further define the runner-up confidence and corresponding confidence margin as
\begin{equation}
\mathrm{second}_j
=
\max_{c\neq\hat{y}_j}p_{j,c},
\qquad
\Delta_j
=
\mathrm{conf}_j-\mathrm{second}_j.
\label{eq:confidence_margin}
\end{equation}
where $\Delta_j$ measures the separation between the two highest class probabilities.


For patch $x_j$, the Prior Synergy Block outputs the tissue-component
probability vector
\begin{equation}
\mathbf{q}_j
=
f_{\mathrm{prior}}(x_j)
=
\left(
q_j^{\mathrm{neu}},
q_j^{\mathrm{sch}},
q_j^{\mathrm{oth}}
\right),
\label{eq:prior_probability}
\end{equation}
where the three entries correspond to the predicted probabilities of neuropil, Schwannian stroma, and other tissue components, respectively. The resulting pathology-prior weight is then computed as
\begin{equation}
w_j^{\mathrm{prior}}
=
\omega_{\mathrm{neu}}q_j^{\mathrm{neu}}
+
\omega_{\mathrm{sch}}q_j^{\mathrm{sch}}
+
\omega_{\mathrm{oth}}q_j^{\mathrm{oth}},
\label{eq:prior_weight}
\end{equation}
where $\omega_{\mathrm{neu}}$, $\omega_{\mathrm{sch}}$, and
$\omega_{\mathrm{oth}}$ denote the tissue-component-specific prior weights.
This probability-weighted formulation retains the uncertainty of the
Prior Synergy Block rather than assigning each patch a single hard
tissue label.

The final voting weight of patch $j$ is defined as

\begin{equation}
w_j
=
\left(\mathrm{conf}_j\right)^{\rho}
\left(1+\lambda\Delta_j\right)
w_j^{\mathrm{prior}},
\label{eq:final_patch_weight}
\end{equation}
where $\rho$ controls the suppression of low-confidence patches and
$\lambda$ controls the contribution of the confidence margin. The term
$\left(\mathrm{conf}_j\right)^{\rho}$ reduces the influence of uncertain
patches, whereas $\left(1+\lambda\Delta_j\right)$ assigns greater weight
to patches with a clearer separation between the two most probable
diagnostic categories. Because $\mathbf{p}_j$ is a probability vector,
$0\leq\Delta_j\leq1$; therefore, for $\lambda\geq0$, the margin-based
amplification factor is bounded within $[1,1+\lambda]$, preventing an
individual patch from receiving an unbounded increase in voting weight.

The coefficients $\omega_{\mathrm{neu}}$, $\omega_{\mathrm{sch}}$, $\omega_{\mathrm{oth}}$, $\rho$, and $\lambda$ are algorithmic hyperparameters that regulate the contributions of tissue-component priors, patch confidence, and confidence margins, rather than clinical diagnostic thresholds. Their values were selected using a held-out validation subset derived from the training cohort and were subsequently fixed without further adjustment for all internal and external test cohorts.

\begin{figure*}[tbp]
    \centering
    \includegraphics[width=\textwidth]{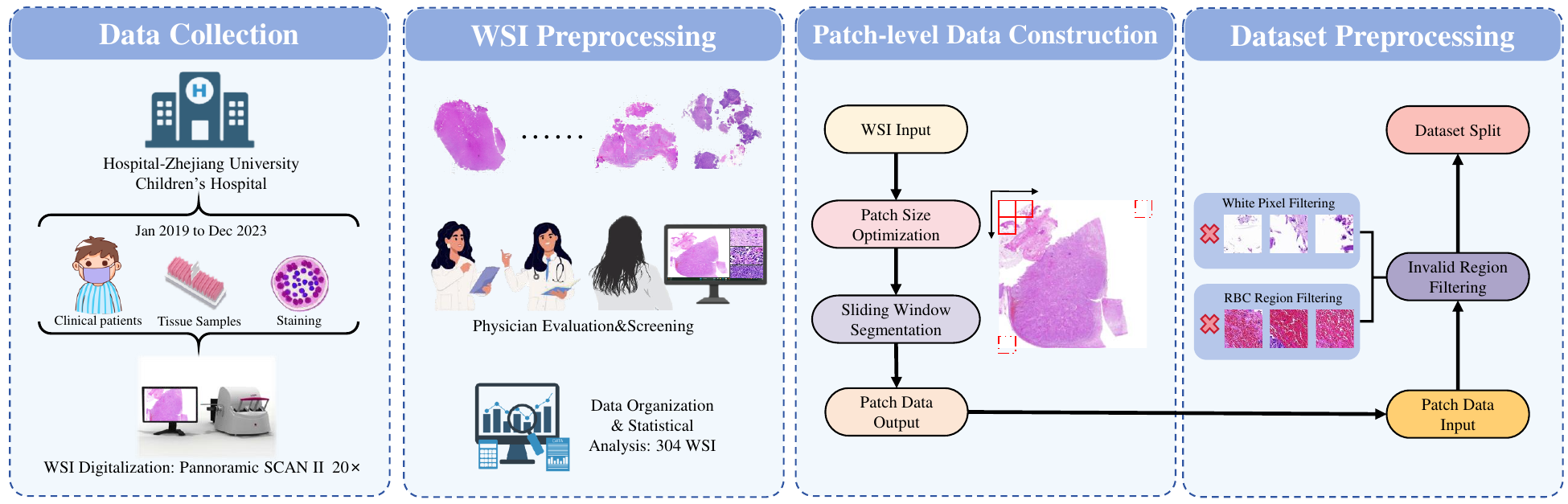} 
    \caption{Processing workflow based on pathological sections of PpNTs, which includes four main stages: (1) Data Collection, covering the collection of clinical patient tissue samples, tissue sample processing, and staining, as well as WSI scanning; (2) WSI preprocessing, involving physician evaluation and screening of WSIs and data organization; (3) Patch-level Data Construction, including determining patch size and sliding window segmentation of slices; (4) Dataset Preprocessing, covering invalid region filtering and dataset splitting,  ultimately providing structured data support for model training and evaluation.}
    \label{fig:data_processing_workflow}
\end{figure*}

\begin{algorithm}[t]
\caption{Pathology-prior-guided WSI voting in PathVote.}
\label{alg:pathvote}
\begin{algorithmic}[1]

\Require Patch-level logits and images
$\{(\mathbf{z}_i,x_i)\}_{i=1}^{n}$, prior model
$f_{\mathrm{prior}}$, and parameters
$(\mathrm{top\_pct},\mathrm{min\_conf},\rho,\lambda,
\tau,\eta,\omega_{\mathrm{neu}},
\omega_{\mathrm{sch}},\omega_{\mathrm{oth}})$

\Ensure WSI-level prediction $y_{\mathrm{pred}}$

\State Compute patch probabilities
$\mathbf{p}_i=\mathrm{softmax}(\mathbf{z}_i)$

\State Compute
$\mathrm{conf}_i=\max_c p_{i,c}$ and
$\hat{y}_i=\arg\max_c p_{i,c}$

\State Set
$\kappa=\max\left(
1,\left\lfloor n\cdot\mathrm{top\_pct}/100\right\rfloor
\right)$

\State Select the Top-$\kappa$ patches ranked by
$\mathrm{conf}_i$ and retain those satisfying
$\mathrm{conf}_i\geq\mathrm{min\_conf}$ to form
$\mathcal{S}$

\For{each selected patch $j\in\mathcal{S}$}

    \State Obtain the tissue-component probabilities
    $\mathbf{q}_j=f_{\mathrm{prior}}(x_j)$

    \State Compute $w_j^{\mathrm{prior}}$ using
    Eq.~\eqref{eq:prior_weight}

    \State Compute $\mathrm{second}_j$ and $\Delta_j$ using
    Eq.~\eqref{eq:confidence_margin}

    \State Compute the final patch weight $w_j$ using
    Eq.~\eqref{eq:final_patch_weight}

\EndFor

\State Aggregate the weighted class logits:
$\mathbf{Z}
=
\sum_{j\in\mathcal{S}}
w_j\mathbf{z}_j$

\State Compute the soft-voting probability:
$\mathbf{p}_{\mathrm{soft}}
=
\mathrm{softmax}(\mathbf{Z}/\tau)$

\State Compute the hard-voting probability:
$p_{\mathrm{hard},c}
=
\frac{
\left|\left\{j\in\mathcal{S}:\hat{y}_j=c\right\}\right|
}{
|\mathcal{S}|
}$

\State Fuse the hard- and soft-voting probabilities:
$\mathbf{p}
=
(1-\eta)\mathbf{p}_{\mathrm{hard}}
+
\eta\mathbf{p}_{\mathrm{soft}}$

\State \Return $y_{\mathrm{pred}}=\arg\max_c p_c$

\end{algorithmic}
\end{algorithm}

\begin{figure*}[tbp]
    \centering
    \includegraphics[width=\textwidth]{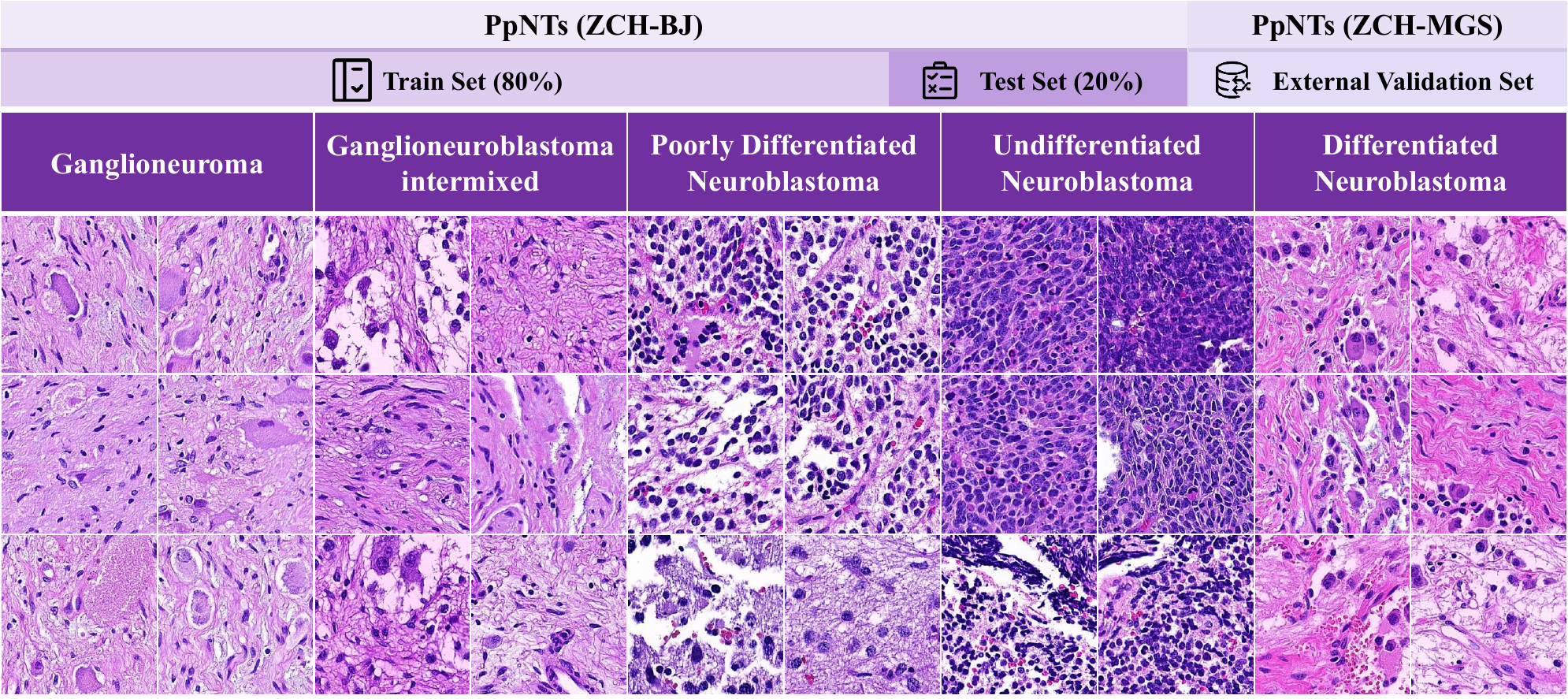} 
    \caption{Visualization of our private PpNTs dataset, which contains five types of neuroblastic tumors: Ganglioneuroma, Ganglioneuroblastoma, Poorly Differentiated Neuroblastoma, Undifferentiated Neuroblastoma, and Differentiated Neuroblastoma.}
    \label{fig:datalist}
\end{figure*}

\section{Experiments}
\subsection{Data Collection and Processing}

Due to the scarcity of pediatric tumors and the complexity of collecting and annotating pathological WSIs, publicly accessible pathological data for pNTs are extremely limited. To address this limitation, we constructed a private two-branch Pediatric Peripheral Neuroblastic Tumor dataset (PpNTs). The dataset consists of 304 WSIs from 102 children diagnosed with pNTs at the Binjiang Branch of Children's Hospital, Zhejiang University School of Medicine (ZCH-BJ), and 109 WSIs from 29 patients diagnosed with pNTs at the Moganshan Branch of the same hospital (ZCH-MGS). It covers five pathological categories: undifferentiated neuroblastoma (UD), poorly differentiated neuroblastoma (PD), differentiated neuroblastoma (D), ganglioneuroblastoma, intermixed (GNBi), and ganglioneuroma (GN). To address intratumoral heterogeneity, only WSIs concordant with the final patient-level diagnosis were retained, and all slides from the same patient were strictly assigned to a single data split to prevent leakage.

The ZCH-BJ cohort was used for training and internal testing, whereas the ZCH-MGS cohort was used as an external validation cohort to evaluate robustness under scanner and magnification shift. WSIs from ZCH-BJ were digitized using a high-precision Pannoramic SCAN II scanner (3DHISTECH, Budapest, Hungary) at $20\times$ magnification. WSIs from ZCH-MGS were scanned at $10\times$ magnification using a Pannoramic MIDI II scanner. This study and dataset were approved by the Academic Ethics Committee of Children's Hospital, Zhejiang University School of Medicine (IRB No. 2023-IRB-0287-P-01), with a waiver of informed consent. Detailed information about the dataset is presented in Fig.~\ref{fig:datalist}, and summary statistics are shown in Table~\ref{tab:dataset_stats}.

Specifically, the private cohort comprised 131 pediatric patients with a median age of 3 years and a relatively balanced sex distribution. Most patients were aged 1-9 years, and all five pathological subtypes were represented across the two branches. Detailed clinical and pathological characteristics are summarized in Table~\ref{tab:clinical_simple}.



\subsubsection{Sliding Window Segmentation}

As shown in Fig. \ref{fig:data_processing_workflow}, a sliding window algorithm is employed with a non-overlapping strategy. The coordinates start from the top-left corner of the WSI (0,0), and the window slides horizontally with a step size of 512 pixels. We discard these parts in the right and bottom areas that cannot be entirely covered by a $512 \times 512$ pixel window, which are primarily blank regions, to avoid introducing invalid information.

\subsubsection{Patch Generation and Quality Filtering}

WSI tiling produces both informative tissue patches and non-informative regions, including blank backgrounds, staining artifacts, and areas densely occupied by red blood cells. To reduce such noise, we apply a two-stage quality-filtering procedure. First, patches in which more than 50\% of pixels have RGB values greater than or equal to 245 are identified as predominantly blank or defective regions and discarded. Second, patches with a red-pixel proportion exceeding 50\% are classified as red-blood-cell-dominant regions and removed because they may obscure diagnostically relevant tissue structures.

\subsubsection{Dataset Construction}
After rigorous screening, the training and test sets are divided based on the pathological category labels of the WSIs in an \(80:20\) ratio. This results in a training set containing \(2{,}379{,}858\) patches and a test set containing \(620{,}862\) patches, ensuring consistent class distribution across both datasets to avoid model performance bias due to sampling variance. All splits were performed at the patient level to prevent cross-set information leakage. The external validation set, PpNTs (ZCH-MGS), containing \(1{,}229{,}563\) patches, is not subjected to dataset splitting.

\begin{table*}[htbp]
  \centering
  \caption{Number of WSIs and patches per category in our private PpNTs dataset.}
  \label{tab:dataset_stats}
  \begin{tabular*}{\linewidth}{@{\extracolsep{\fill}} l l *{12}{c} }
    \toprule
    \multirow{2.7}{*}{\textbf{Dataset}} & \multirow{2.7}{*}{\textbf{Class}} & \multicolumn{2}{c}{\textbf{UD}} & \multicolumn{2}{c}{\textbf{PD}} & \multicolumn{2}{c}{\textbf{D}} & \multicolumn{2}{c}{\textbf{GNBi}} & \multicolumn{2}{c}{\textbf{GN}} & \multicolumn{2}{c}{\textbf{Total}} \\
    \cmidrule(lr){3-4} \cmidrule(lr){5-6} \cmidrule(lr){7-8} \cmidrule(lr){9-10} \cmidrule(lr){11-12} \cmidrule(lr){13-14}
    & & WSI & Patch & WSI & Patch & WSI & Patch & WSI & Patch & WSI & Patch & WSI & Patch\\
    \midrule
    \multirow{2}{*}{\textbf{Internal dataset}} & Train & 50 & 451,986 & 44 & 432,627  & 49 & 474,440 & 50 & 484,792 & 48 & 536,013 & 241 &  2,379,858 \\
    & Test & 13 & 103,438 & 12 & 116,903 & 13 & 151,007 & 13 & 138,995 & 12 & 110,519 & 63 & 620,862 \\
    \midrule
    \textbf{External validation set} & Ext-Val  & 13 & 123,501 & 25 & 270,021 & 36 & 421,493 & 26 & 320,706 & 9 & 93,842 & 109 & 1,229,563 \\
    \bottomrule
  \end{tabular*}
\end{table*}

\begin{table*}[htbp]
  \centering
  \caption{Clinical and pathological characteristics of the 131 pediatric patients with peripheral neuroblastic tumors in the private cohort.}
  \label{tab:clinical_simple}
  \begin{tabular*}{\linewidth}{@{\extracolsep{\fill}}lcccccccccc@{}}
    \toprule
    Age Group & Male & Female & UD & PD & D & GNBi & GN & ZCH-BJ & ZCH-MGS & Count \\
    \midrule
    $0 \leq \text{Age} < 1$   & 4  & 4  & 0 & 5 & 3 & 0 & 0 & 6  & 2 & 8   \\
    $1 \leq \text{Age} < 5$   & 35  & 28  & 16 & 13 & 19 & 11 & 4 & 49  & 14 & 63  \\
    $5 \leq \text{Age} < 10$   & 23  & 22  & 9 & 8 & 3 & 10 & 15 & 36  & 9 & 45  \\
    $10 \leq \text{Age} < 15$   & 8 & 6 & 0 & 0 & 2 & 6 & 6 & 10 & 4 & 14  \\
    $15 \leq \text{Age} < 18$   & 1 & 0  & 0 & 0 & 1 & 0 & 0 & 1 & 0 & 1  \\
    \midrule
    Total                     & 71 & 60 & 25 & 26 & 28 & 27 & 25 & 102 & 29 & 131 \\
    \bottomrule
  \end{tabular*}
\end{table*}

\begin{table*}[htbp]
    \centering
    \caption{Settings, parameters, and comparisons with the baseline of different versions of CoHisNet models.}
    \label{tab3}
    \begin{tabularx}{\textwidth}{
        >{\raggedright\arraybackslash}p{3.0cm} 
        >{\centering\arraybackslash}X 
        >{\centering\arraybackslash}X 
        >{\centering\arraybackslash}X 
        >{\centering\arraybackslash}X 
        >{\centering\arraybackslash}X
    }
    \toprule
    \multirow{2}{*}{\textbf{Model}} & \textbf{Embedding Dimension} & \multirow{2}{*}{\textbf{Depths}} & \multirow{2}{*}{\textbf{Num of Heads}} & \multirow{2}{*}{\textbf{Params}} & \multirow{2}{*}{\textbf{FLOPs}} \\
    \midrule
    Swin Transformer~\cite{swintransformer} & 128 & (2, 2, 18, 2) & (4, 8, 16, 32) & 86.75 M & 15.17 G\\
    \textbf{CoHisNet-mini} & 24 & (2, 2, 8, 2) & (2, 4, 8, 16) & 1.29 M & 0.21 G \\
    \textbf{CoHisNet-micro} & 24 & (2, 2, 8, 2) & (4, 8, 16, 32) & 1.32 M & 0.22 G \\
    \textbf{CoHisNet-tiny} & 56 & (2, 2, 8, 2) & (4, 8, 16, 32) & 5.52 M & 1.08 G \\
    \bottomrule
    \end{tabularx}
\end{table*}

\subsection{Public Dataset}
We further evaluated CoHisNet on the public BreakHis dataset~\cite{spanhol2015dataset}, which contains 7,909 breast histopathology images from 82 patients across eight tumor categories and four magnifications. We used the 1,820 images acquired at $\times400$ magnification. The original training partition contained 1,427 images and was augmented to 6,109 samples to mitigate class imbalance, whereas the test partition contained 393 images.

\subsection{Experimental Setup}


All experiments were performed on an NVIDIA Tesla V100-SXM2 GPU with 32 GB memory, under a fixed software environment using Python 3.8.20 and PyTorch 2.4.1 with CUDA 12.1. Models were trained for 30 epochs with a batch size of 64, requiring approximately 75 hours on the private dataset. Input patches were resized to $224\times224$ pixels and normalized with ImageNet statistics. CoHisNet was initialized with He initialization~\cite{2015Delving} and optimized by AdamW with a cosine annealing learning-rate schedule. For PathVote, the tissue-component-specific prior weights were set to $\omega_{\mathrm{neu}}=1$, $\omega_{\mathrm{sch}}=8$, and $\omega_{\mathrm{oth}}=1$. Detailed model configurations are reported in Table~\ref{tab3}.

For fair comparison, all trainable non-foundation baselines followed the same protocol as CoHisNet, including patient-level splitting, patch extraction, preprocessing, optimization, and training schedule. For foundation models, including UNI, UNI2, CTransPath, CONCH, and KEEP, the pretrained backbone was frozen, and only a task-specific linear classifier was trained under the same downstream split and preprocessing pipeline. In WSI-level evaluation, all models used identical patch-level inference and slide-level aggregation protocols.

All KAN layers were optimized jointly with the backbone using AdamW and the same cosine-annealing schedule, without layer-specific learning rates or gradient clipping. We used a memory-efficient implementation that vectorizes B-spline basis evaluation and dense projection, avoiding explicit sample-wise edge-activation expansion. The predefined knot grid remained fixed throughout training, and no specialized sparse kernel was used.

Statistical significance was assessed using patient-clustered paired bootstrap resampling, with all samples from the same patient retained within each resampled cluster. For each metric, the best-performing CoHisNet variant was compared with the best-performing non-CoHisNet baseline. A two-sided $p<0.05$ was considered statistically significant.

\begin{table*}[t]
    \centering
    \caption{Comparison of CoHisNet with other advanced models on our private PpNTs dataset.}
    \label{tabpatch}
    \setlength{\tabcolsep}{1pt}
    \begin{tabularx}{\textwidth}{
        >{\raggedright\arraybackslash}p{3cm}
        >{\centering\arraybackslash}X 
        >{\centering\arraybackslash}X 
        >{\centering\arraybackslash}X 
        >{\centering\arraybackslash}X 
        >{\centering\arraybackslash}X 
        >{\centering\arraybackslash}X 
        >{\centering\arraybackslash}X 
        >{\centering\arraybackslash}X 
        >{\centering\arraybackslash}X 
        >{\centering\arraybackslash}X 
    }
    \toprule
    \textbf{Dataset} & \multicolumn{5}{c}{\textbf{Internal test set}} & \multicolumn{5}{c}{\textbf{External validation set}}\\
    \cmidrule(lr){2-6} \cmidrule(lr){7-11}
    \textbf{Model} & \textbf{ACC} & \textbf{BACC} & \textbf{Kappa} & \textbf{F1} & \textbf{AUROC} & \textbf{ACC} & \textbf{BACC} & \textbf{Kappa} & \textbf{F1} & \textbf{AUROC} \\
    \midrule
    \multicolumn{11}{c}{General method} \\
    \midrule
    VGGNet \cite{vgg} & 81.89 & 81.52 & 74.37 & 81.76 & 97.02 & 28.83 & 26.78 & 9.363 & 25.98 & 50.31 \\
    ResNet-50 \cite{resnet} & 92.75 & 92.35 & 90.94 & 92.73 & 99.43 & 46.91 & 40.68 & 29.76 & 47.83 & 71.35 \\
    DenseNet \cite{densnet} & 92.02 & 91.73 & 89.71 & 91.93 & 99.36 & 48.70 & 41.20 & 33.17 & 49.71 & 75.79 \\
    Vision Transformer \cite{visiontransformer} & 89.24 & 89.12 & 85.90 & 89.23 & 98.79 & 39.74 & 34.78 & 28.56 & 47.41 & 58.73 \\
    Swin Transformer \cite{swintransformer} & 91.81 & 91.68 & 89.64 & 91.81 & 99.30 & 47.64 & 41.80 & 30.94 & 48.82 & 73.76 \\
    \midrule
    \multicolumn{11}{c}{Pathological large-scale pre-trained model} \\
    \midrule
    UNI \cite{chen2024towards} & 93.35 & 93.12 & 91.64 & 93.29 & 99.43 & 49.91 & 45.33 & 19.04 & 50.09 & 77.17 \\
    UNI2 \cite{chen2024towards} & 93.19 & 93.00 & 91.43 & 93.18 & 99.14 & 51.75 & 46.98 & 21.85 & 51.74 & 77.52\\
    CTransPath \cite{wang2022transformer} & 92.22 & 92.11 & 90.22 & 92.24 & 99.36 & 50.91 & 40.13 & 37.01 & \textbf{54.37} & 79.81 \\
    CONCH \cite{lu2024visual} & 92.17 & 92.01 & 90.32 & 92.15 & 95.66 & 49.17 & 43.81 & 24.12 & 50.02 & 66.04 \\
    KEEP \cite{ZHOU2026777} & 93.02 & 92.84 & 91.21 & 93.00 & 99.24 & 49.62 & 45.12 & 34.72 & 49.82 & 78.29\\
    \midrule
    \multicolumn{11}{c}{Pathological classification model} \\
    \midrule
    HisImage \cite{ding2023enhanced} & 86.29 & 86.26 & 82.78 & 86.21 & 98.00 & 46.49 & 41.98 & 30.52 & 40.33 & 73.81 \\
    FMDNN \cite{ding2024fmdnn} & 92.53 & 92.30 & 89.81 & 92.48 & 99.41 & 45.94 & 41.07 & 20.51 & 47.03 & 75.52 \\
    StoHisNet \cite{fu2022stohisnet} & 92.21 & 92.03 & 90.87 & 92.18 & 99.33 & 21.26 & 20.62 & 1.214 & 9.257 & 47.52 \\
    HiFuse \cite{huo2023hifuse} & 91.49 & 91.40 & 89.31 & 91.49 & 99.29 & 51.47 & 42.09 & 34.98 & 51.42 & 78.82 \\
    \midrule
    \multicolumn{11}{c}{Ours} \\
    \midrule
    \textbf{CoHisNet-mini} & 93.37 & 93.18 & 91.77 & 93.36 & 99.46 & 49.40 & 41.61 & 35.34 & \underline{53.68} & \underline{79.98} \\
    \textbf{CoHisNet-micro} & \textbf{93.58} & \textbf{93.42} & \textbf{92.12} & \textbf{93.57} & \textbf{99.53} & \underline{51.86} & \textbf{48.02}\textsuperscript{*} & \underline{37.81} & 51.72 & 77.85 \\
    \textbf{CoHisNet-tiny} & \underline{93.43} & \underline{93.52} & \underline{91.96} & \underline{93.42} & \underline{99.51} & \textbf{52.86} & \underline{47.63} & \textbf{38.29} & 52.28 & \textbf{83.57}\textsuperscript{*} \\
    \bottomrule
    \addlinespace[2pt]
    \multicolumn{11}{r}{\footnotesize $^{*}p<0.05$ versus the best-performing non-CoHisNet baseline for the corresponding metric.} \\
    \end{tabularx}
\end{table*}

\begin{table}[htbp]
    \centering
    \caption{Comparison of CoHisNet with other advanced models on the public BreakHis dataset.}
    \label{tab_out}
    \begin{tabularx}{\columnwidth}{
        >{\raggedright\arraybackslash}p{2.8cm}
        >{\centering\arraybackslash}X 
        >{\centering\arraybackslash}X 
        >{\centering\arraybackslash}X 
        >{\centering\arraybackslash}X 
        >{\centering\arraybackslash}X 
    }
    \toprule
    \textbf{Model} & \textbf{ACC} & \textbf{BACC} & \textbf{Kappa} & \textbf{F1} & \textbf{AUROC} \\
    \midrule
    \multicolumn{6}{c}{General method} \\
    \midrule
    VGGNet \cite{vgg} & 48.09 & 23.57 & 21.44 & 41.37 & 67.71 \\
    ResNet-50 \cite{resnet} & 68.45 & 55.61 & 47.65 & 67.40 & 92.37 \\
    DenseNet \cite{densnet} & 83.21 & 80.81 & 77.33 & 83.64 & 98.25 \\
    Vision Transformer \cite{visiontransformer} & 68.45 & 59.84 & 61.27 & 67.58 & 91.03 \\
    Swin Transformer \cite{swintransformer} & 75.82 & 71.22 & 68.38 & 76.06 & 96.37 \\
    \midrule
    \multicolumn{6}{c}{Pathological large-scale pre-trained model} \\
    \midrule
    UNI \cite{chen2024towards} & 78.88 & 69.91 & 72.07 & 71.41 & 97.03 \\
    UNI2 \cite{chen2024towards} & 80.92 & 72.09 & 74.69 & 73.52 & 98.17 \\
    CTransPath \cite{wang2022transformer} & 83.46 & 86.62 & 78.92 & 83.79 & 98.52 \\
    CONCH \cite{lu2024visual} & \underline{86.77} & 86.99 & \textbf{86.98} & 86.81 & 94.73 \\
    KEEP \cite{ZHOU2026777} & 65.39 & 62.01 & 53.97 & 59.77 & 92.76 \\
    \midrule
    \multicolumn{6}{c}{Pathological classification model} \\
    \midrule
    HisImage \cite{ding2023enhanced} & 64.63 & 55.96 & 52.50 & 57.59 & 84.43 \\
    FMDNN \cite{ding2024fmdnn} & 86.32 & 81.68 & 84.71 & 86.14 & 96.57 \\
    StoHisNet \cite{fu2022stohisnet} & 74.84 & 68.33 & 67.32 & 74.24 & 94.26 \\
    HiFuse \cite{huo2023hifuse} & 62.84 & 47.93 & 48.78 & 60.59 & 87.02 \\
    \midrule
    \multicolumn{6}{c}{Ours} \\
    \midrule
    \textbf{CoHisNet-mini} & 84.98 & 84.48 & 81.31 & 85.12 & 98.61 \\
    \textbf{CoHisNet-micro} & 86.51 & \underline{87.04} & 83.87 & \underline{86.90} & \underline{98.64} \\
    \textbf{CoHisNet-tiny} & \textbf{87.79} & \textbf{87.56} & \underline{85.69} & \textbf{88.04} & \textbf{98.70} \\
    \bottomrule
    \end{tabularx}
\end{table}

\begin{table*}[tbp]
    \centering
    \caption{Comparison of CoHisNet with other advanced models with PathVote on our private PpNTs dataset.}
    \label{tab6}
    \begin{tabularx}{\textwidth}{
        >{\raggedright\arraybackslash}p{4.2cm}
        >{\centering\arraybackslash}X 
        >{\centering\arraybackslash}X 
        >{\centering\arraybackslash}X 
        >{\centering\arraybackslash}X 
        >{\centering\arraybackslash}X 
        >{\centering\arraybackslash}X 
        >{\centering\arraybackslash}X 
        >{\centering\arraybackslash}X 
    }
    \toprule
    \textbf{Dataset} & \multicolumn{4}{c}{\textbf{Internal test set}} & \multicolumn{4}{c}{\textbf{External validation set}}\\
    \cmidrule(lr){2-5} \cmidrule(lr){6-9}
    \textbf{Model} & \textbf{ACC} & \textbf{BACC} & \textbf{Kappa} & \textbf{F1} & \textbf{ACC} & \textbf{BACC} & \textbf{Kappa} & \textbf{F1} \\
    \midrule
    \multicolumn{9}{c}{General method} \\
    \midrule
    VGGNet \cite{vgg} (hard vote) & 92.06 & 91.79 & 90.07 & 91.88 & 48.67 & 52.88 & 35.15 & 49.04 \\
    VGGNet \cite{vgg} (PathVote) & 93.65 & 93.46 & 92.06 & 93.55 & 56.88 & 53.40 & 44.53 & 58.23 \\
    ResNet-50 \cite{resnet} (hard vote) & 98.41 & 98.33 & 98.01 & 98.39 & 69.72 & 54.70 & 58.91 & 52.21 \\
    ResNet-50 \cite{resnet} (PathVote) & 98.41 & 98.33 & 98.01 & 98.39 & 72.48 & 64.59 & 63.88 & 72.62 \\
    DenseNet \cite{densnet} (hard vote) & 98.41 & 98.33 & 98.01 & 98.39 & 74.31 & 64.20 & 65.19 & 66.37 \\
    DenseNet \cite{densnet} (PathVote) & 98.41 & 98.33 & 98.01 & 98.39 & 75.23 & 69.92 & 67.38 & 75.21 \\
    Vision Transformer \cite{visiontransformer} (hard vote) & 98.41 & 98.33 & 98.01 & 98.39 & 60.68 & 55.24 & 48.76 & 60.83 \\
    Vision Transformer \cite{visiontransformer} (PathVote) & 98.41 & 98.33 & 98.01 & 98.39  & 64.22 & 64.91 & 54.21 & 65.27 \\
    Swin Transformer \cite{swintransformer} (hard vote) & 98.41 & 98.33 & 98.01 & 98.39 & 68.81 & 55.44 & 57.49 & 53.62 \\
    Swin Transformer \cite{swintransformer} (PathVote) & 98.41 & 98.33 & 98.01 & 98.39 & 71.56 & 68.16 & 63.25 & 71.97 \\
    \midrule
    \multicolumn{9}{c}{Pathological large-scale pre-trained model} \\
    \midrule
    UNI \cite{chen2024towards} (hard vote) & 98.41 & 98.33 & 98.01 & 98.39 & 75.23 & 68.85 & 67.35 & 69.34 \\
    UNI \cite{chen2024towards} (PathVote) & 98.41 & 98.33 & 98.01 & 98.39  & 76.15 & 69.80 & 68.67 & 70.02 \\
    UNI2 \cite{chen2024towards} (hard vote) & 98.41 & 98.33 & 98.01 & 98.39 & 73.39 & 66.54 & 65.00 & 66.24 \\
    UNI2 \cite{chen2024towards} (PathVote) & 98.41 & 98.33 & 98.01 & 98.39  & 74.31 & 68.08 & 66.17 & 67.83 \\
    CTransPath \cite{wang2022transformer} (hard vote) & 98.41 & 98.33 & 98.01 & 98.39 & 77.98 & 72.59 & 71.07 & 78.00 \\
    CTransPath \cite{wang2022transformer} (PathVote) & 98.41 & 98.33 & 98.01 & 98.39 & 79.82 & 75.04 & 73.53 & 79.90 \\
    CONCH \cite{lu2024visual} (hard vote) & 98.41 & 98.33 & 98.01 & 98.39 & 71.56 & 66.71 & 62.86 & 72.09 \\
    CONCH \cite{lu2024visual} (PathVote) & 98.41 & 98.33 & 98.01 & 98.39 & 74.31 & 69.12 & 65.94 & 73.96 \\
    KEEP \cite{ZHOU2026777} (hard vote) & 98.41 & 98.33 & 98.01 & 98.39 & 72.48 & 65.71 & 63.84 & 65.69 \\
    KEEP \cite{ZHOU2026777} (PathVote) & 98.41 & 98.33 & 98.01 & 98.39 & 75.23 & 68.85 & 67.43 & 69.38 \\

    \midrule
    \multicolumn{9}{c}{Pathological classification model} \\
    \midrule
    HisImage \cite{ding2023enhanced} (hard vote) & 95.24 & 95.13 & 94.04 & 95.20 & 52.29 & 45.69 & 38.56 & 44.26 \\
    HisImage \cite{ding2023enhanced} (PathVote) & 96.83 & 96.79 & 96.03 & 96.82 & 58.72 & 49.94 & 45.82 & 49.34 \\
    FMDNN \cite{ding2024fmdnn} (hard vote) & 98.41 & 98.33 & 98.01 & 98.39 & 53.21 & 43.26 & 38.02 & 52.28 \\
    FMDNN \cite{ding2024fmdnn} (PathVote) & 98.41 & 98.33 & 98.01 & 98.39 & 58.72 & 51.19 & 45.63 & 80.63 \\
    StoHisNet \cite{fu2022stohisnet} (hard vote) & 98.41 & 98.33 & 98.01 & 98.39 & 37.61 & 31.36 & 18.73 & 25.93 \\
    StoHisNet \cite{fu2022stohisnet} (PathVote) & 98.41 & 98.33 & 98.01 & 98.39 & 41.28 & 34.04 & 23.01 & 71.37 \\
    HiFuse \cite{huo2023hifuse} (hard vote) & 98.41 & 98.33 & 98.01 & 98.39 & 67.89 & 53.75 & 56.22 & 52.50 \\
    HiFuse \cite{huo2023hifuse} (PathVote) & 98.41 & 98.33 & 98.01 & 98.39 & 68.81 & 55.01 & 57.56 & 53.75 \\
    \midrule
    \multicolumn{9}{c}{Ours} \\
    \midrule
    \textbf{CoHisNet-mini (hard vote)} & 98.41 & 98.33 & 98.01 & 98.39 & 78.90 & 72.36 & 71.99 & 78.32 \\
    \textbf{CoHisNet-mini (PathVote)} & 98.41 & 98.33 & 98.01 & 98.39 & 79.82 & 77.10 & 73.13 & 79.00 \\
    \textbf{CoHisNet-micro (hard vote)} & 98.41 & 98.33 & 98.01 & 98.39 & 80.73 & 77.48 & 74.58 & 80.68 \\
    \textbf{CoHisNet-micro (PathVote)} &\textbf{100.00} &\textbf{100.00} &\textbf{100.00} &\textbf{100.00} & 82.57 & \underline{78.39} & 77.17 &\textbf{82.54} \\
    \textbf{CoHisNet-tiny (hard vote)} & 98.41 & 98.33 & 98.01 & 98.39 & \underline{83.49} & 78.25 & \underline{77.95} & 80.74 \\
    \textbf{CoHisNet-tiny (PathVote)} & 98.41 & 98.33 & 98.01 & 98.39 & \textbf{84.40}  & \textbf{79.73}\textsuperscript{*}  & \textbf{79.17}\textsuperscript{*}  & \underline{82.38} \\
    
    \bottomrule
    \addlinespace[2pt]
    \multicolumn{9}{r}{\footnotesize $^{*}p<0.05$ versus the best-performing non-CoHisNet baseline for the corresponding metric.} \\
    \end{tabularx}
\end{table*}

\subsection{Comparative Experiments}
\subsubsection{Model Performance Comparison}

\begin{figure}[h]
    \centering
    \includegraphics[width=\columnwidth]{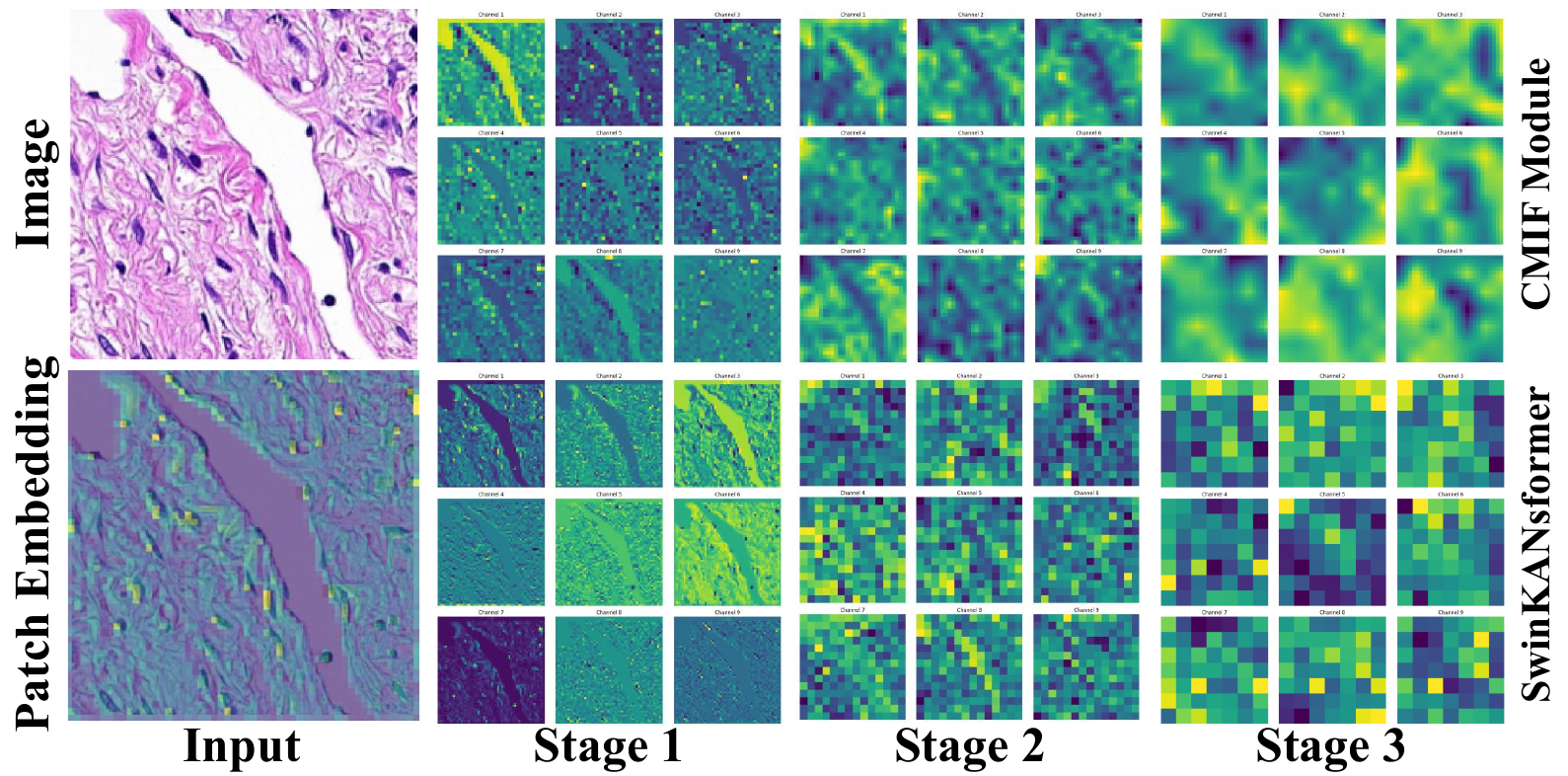} 
\caption{Feature maps produced by the Swin KANsformer stages and the CMIF module for representative H\&E-stained patches. The visualization illustrates the evolution of spatial responses across hierarchical stages and multi-scale fusion.}
    \label{fig:att}
\end{figure}

As shown in Table~\ref{tabpatch}, CoHisNet-micro achieves the highest ACC on the internal PpNTs (ZCH-BJ) test set, exceeding Swin Transformer and Vision Transformer by 1.77 and 4.34 percentage points, respectively. The stage-wise feature responses in Fig.~\ref{fig:att} illustrate how CMIF and KAN progressively refine fine-grained morphology and high-level semantic information within individual patches for patch-level classification.

Compared with pathology foundation models, our model, which uses random initialization without pre-training, still outperformed CONCH and UNI, which were pre-trained on large-scale pathological datasets and transferred to our private dataset. While these models perform well in multimodal pathological data, their effectiveness in capturing subtle features and differentiating morphologically similar histopathological patterns in specialized fields, such as pNT subtype classification, is limited.

We compared CoHisNet with advanced pathology-specific classification models such as HisImage \cite{ding2023enhanced}, FMDNN \cite{ding2024fmdnn}, and HiFuse \cite{huo2023hifuse}, which struggle with similar categorical features. Under the same training conditions, CoHisNet-micro surpasses them in every metric, achieving a Kappa of 92.12\% versus their 82.78\%, 89.81\%, and 89.31\%. Compared to the improved baseline models CTransPath \cite{wang2022transformer} and StoHisNet \cite{fu2022stohisnet}, CoHisNet-micro also improved ACC by 1.36\% and 1.37\%, respectively. In the external validation set, PpNTs (ZCH-MGS), which is critical for evaluating the generalizability, CoHisNet also showed consistent superiority.

\begin{figure}[t]
    \centering
    \includegraphics[width=\columnwidth]{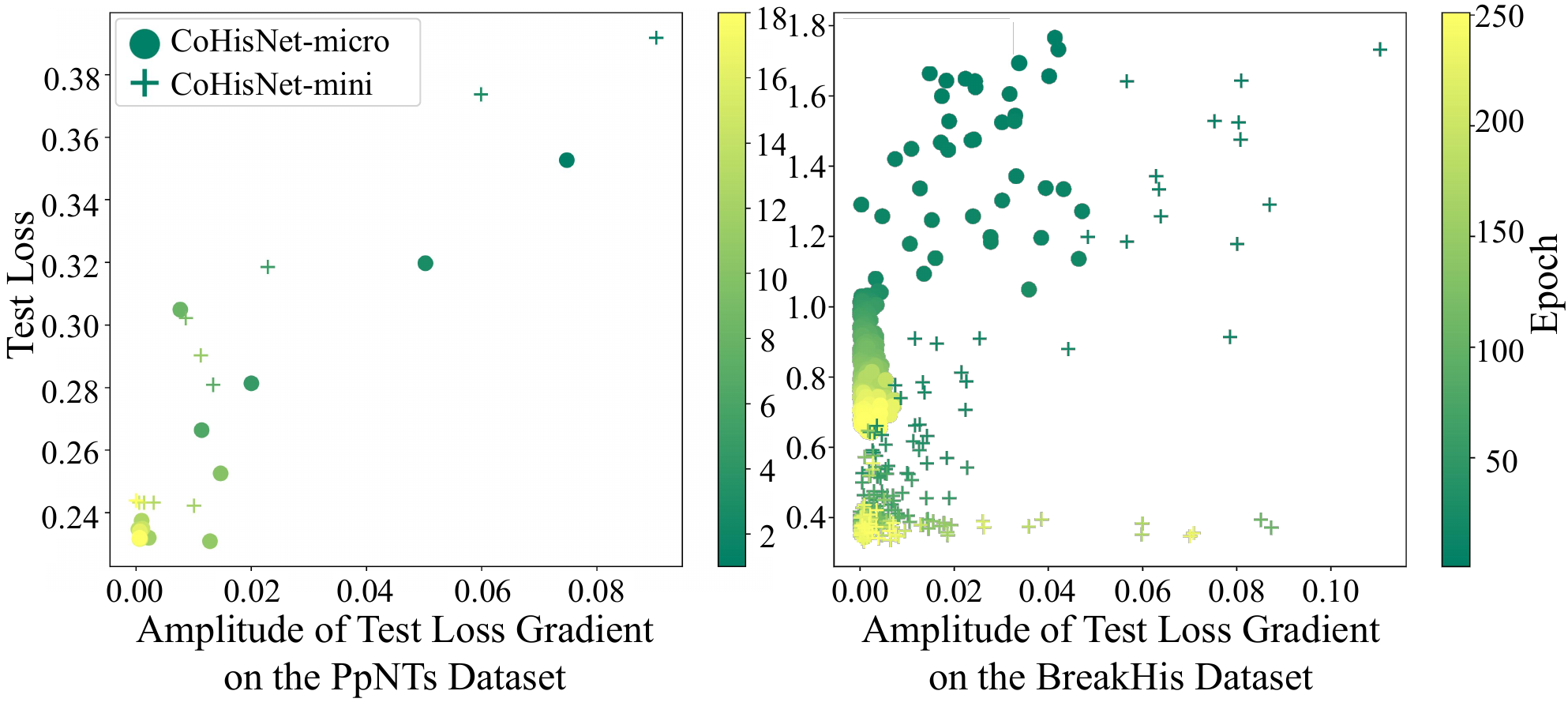} 
    \caption{The loss-gradient plots for different versions of the CoHisNet models evaluated on PpNTs and BreakHis datasets.}
    \label{fig:loss}
\end{figure}

\begin{figure}[t]
    \centering
    \includegraphics[width=\columnwidth]{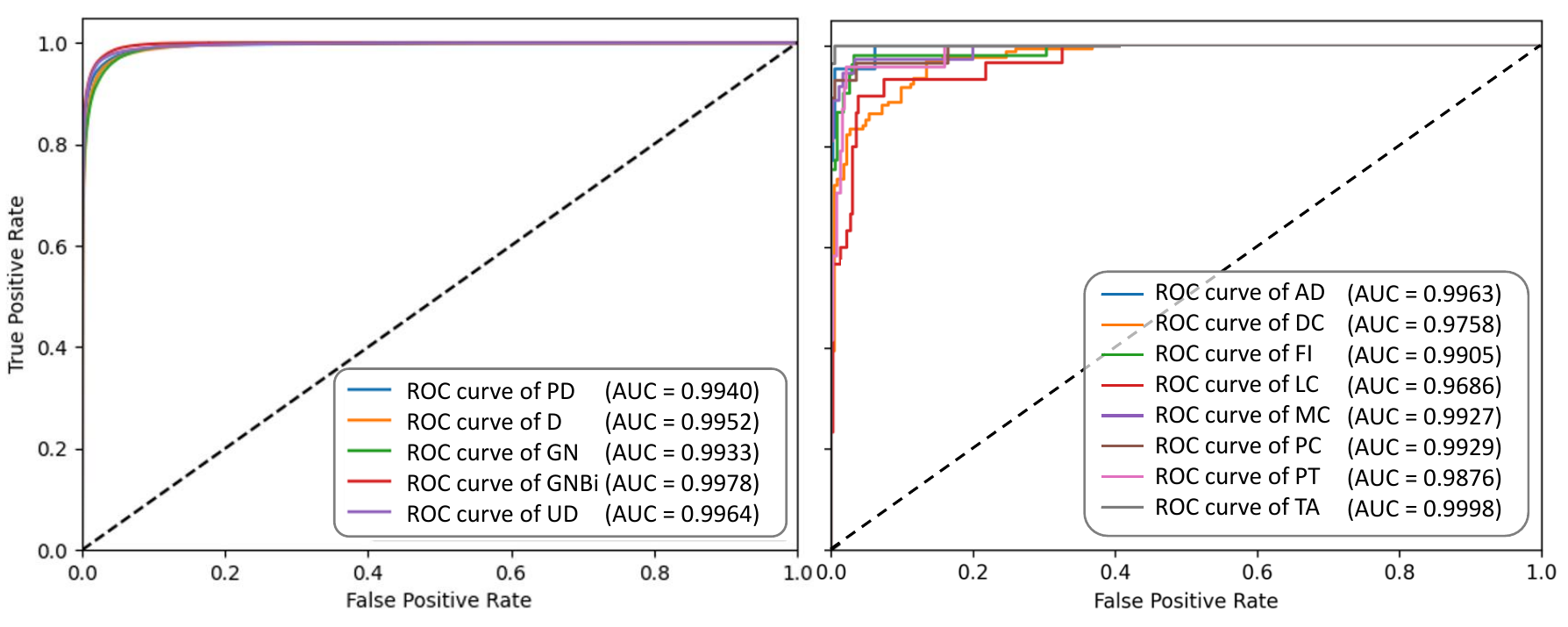} 
    \caption{ROC curves of CoHisNet on our private PpNTs and public BreakHis test sets.}
    \label{fig:roc}
\end{figure}

\begin{figure*}[htbp]
    \centering
    \includegraphics[width=\textwidth]{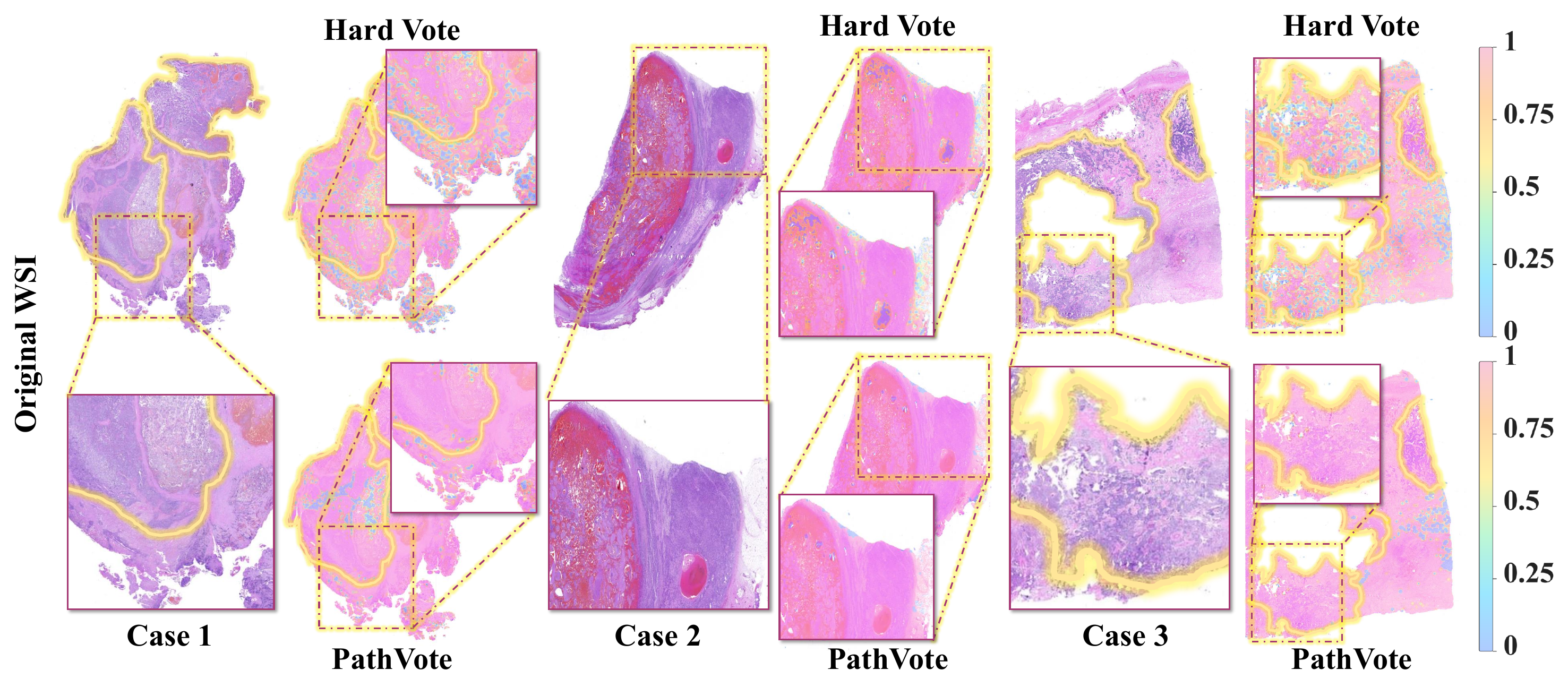} 
\caption{Representative WSI-level heatmaps obtained with hard voting and PathVote. Yellow contours delineate pathologist-annotated diagnostic regions. Low values (blue) correspond to incorrectly classified patches, which provide misleading evidence during WSI-level aggregation. Compared with hard voting, PathVote yields fewer low-value regions and preserves stronger responses within the annotated regions.}

    \label{fig:wsi_heatmap}
\end{figure*}

In the WSI-level experiments with PathVote, as shown in Table~\ref{tab6}, CoHisNet-micro achieved 100.00\% ACC and 100.00\% BACC on the internal test set. This result should be interpreted in the context of the strong patch-level performance on the same cohort. Because WSI-level prediction aggregates many patch-level outputs, isolated patch-level errors can be diluted when most diagnostically informative patches are correctly classified with high confidence. Therefore, high patch-level accuracy may translate into perfect WSI-level accuracy on a relatively small internal test set. In our internal WSI-level test set of 63 WSIs, one misclassified case would reduce the overall accuracy from 100.00\% to 98.41\%. Thus, the internal WSI-level result should not be regarded as evidence of error-free diagnosis, but rather as an indication that PathVote can effectively aggregate reliable patch-level predictions. The external validation cohort provides a more informative assessment of robustness under scanner and magnification shift. Fig.~\ref{fig:wsi_heatmap} shows representative WSI-level heatmaps, where blue regions indicate erroneous or low-reliability patch predictions that may adversely affect slide-level voting.

On the external validation cohort PpNTs (ZCH-MGS), CoHisNet-tiny with PathVote achieved 84.40\% ACC. This was 9.17 percentage points higher than the best general model with PathVote, DenseNet, and 12.84 percentage points higher than Swin Transformer. It also exceeded CTransPath by 4.58 percentage points and HiFuse by 15.59 percentage points. These results indicate that PathVote can improve WSI-level robustness when tissue-component priors are applicable to the target pathology by stabilizing subtype-discriminative evidence despite acquisition-related appearance variation and heterogeneous tissue composition.

Within the CoHisNet family, PathVote consistently outperformed hard voting. For CoHisNet-tiny, PathVote improved BACC from 78.25\% to 79.73\% and F1 from 80.74\% to 82.38\% on the external validation cohort. These results support the use of probabilistic, pathology-informed fusion rather than equal-weight majority voting.

\subsubsection{Validation of Model Generalization Ability}




To further evaluate cross-dataset generalization, we compared CoHisNet with the same three categories of models on the public BreakHis dataset, as shown in Table~\ref{tab_out}. Among general image classifiers, Swin Transformer achieved 75.82\% ACC, whereas CoHisNet-tiny achieved 87.79\% ACC. Compared with pathology foundation models, CoHisNet-tiny achieved higher ACC than CTransPath and UNI, while CoHisNet-micro achieved slightly higher BACC than CTransPath. Among pathology-specific classifiers, CoHisNet-tiny also achieved the highest ACC and F1. These results suggest that the proposed multi-scale interaction and KAN-based nonlinear modeling are not limited to pNTs, although BreakHis differs from PpNTs in tumor type, imaging setting, and task definition. Figs.~\ref{fig:loss} and~\ref{fig:roc} present loss gradients and ROC curves for both datasets.


\begin{table}[tbp]
    \centering
    \caption{Comparison of CoHisNet with other advanced models with hard vote on the BreakHis dataset.}
    \label{tab7}
    \begin{tabularx}{\columnwidth}{
        >{\raggedright\arraybackslash}p{4.2cm}
        >{\centering\arraybackslash}X 
        >{\centering\arraybackslash}X 
        >{\centering\arraybackslash}X 
        >{\centering\arraybackslash}X 
    }
    \toprule
    \textbf{Model} & \textbf{ACC} & \textbf{BACC} & \textbf{Kappa} & \textbf{F1}  \\
    \midrule
    \multicolumn{5}{c}{General method} \\
    \midrule
    VGGNet \cite{vgg} (hard vote) & 46.91 & 33.45 & 32.00 & 23.60 \\
    ResNet-50 \cite{resnet} (hard vote) & 60.49 & 63.36 & 51.98 & 54.98 \\
    DenseNet \cite{densnet} (hard vote) & 93.83 & 94.21 & 91.75 & 92.67 \\
    Vision Transformer \cite{visiontransformer} (hard vote) & 66.67 & 60.28 & 55.50 & 57.88 \\
    Swin Transformer \cite{swintransformer} (hard vote) & 80.25 & 75.54 & 73.80 & 73.71 \\
    \midrule
    \multicolumn{5}{c}{Pathological large-scale pre-trained model} \\
    \midrule
    UNI \cite{chen2024towards} (hard vote) & 88.89 & 78.99 & 84.70 & 82.49 \\
    UNI2 \cite{chen2024towards} (hard vote) & 90.12 & 86.97 & 86.67 & 90.21 \\
    CTransPath \cite{wang2022transformer} (hard vote) & 92.59 &94.56 & 90.17 & 91.84 \\
    CONCH \cite{lu2024visual} (hard vote) &\underline{95.05} & 93.75 & 93.28 & 94.51 \\
    KEEP \cite{ZHOU2026777} (hard vote) &74.39 & 72.02 & 65.52 & 69.26 \\
    \midrule
    \multicolumn{5}{c}{Pathological classification model} \\
    \midrule
    HisImage \cite{ding2023enhanced} (hard vote) & 75.31 & 69.36 & 65.60 & 69.26 \\
    FMDNN \cite{ding2024fmdnn} (hard vote) & 92.59 & 93.73 & 90.17 & 91.37 \\
    StoHisNet \cite{fu2022stohisnet} (hard vote) & 85.23 & 75.84 & 67.09 & 84.23 \\
    HiFuse \cite{huo2023hifuse} (hard vote) & 64.20 & 43.76 & 47.43 & 43.80 \\
    \midrule
    \multicolumn{5}{c}{Ours} \\
    \midrule
    \textbf{CoHisNet-mini (hard vote)} & 93.83 & 94.32 & 91.70 & 93.84\\ 
    \textbf{CoHisNet-micro (hard vote)} & \textbf{95.06} & \underline{94.66} & \underline{93.32} & \underline{94.99}\\
    \textbf{CoHisNet-tiny (hard vote)} &\textbf{95.06} &\textbf{96.69} &\textbf{93.40} &\textbf{95.34} \\
    \bottomrule
    \end{tabularx}
\end{table}

We conducted WSI classification experiments on the BreakHis dataset to further confirm its generalization ability. Because BreakHis lacks the specific clinical priors required by PathVote, we could not employ PathVote. As shown in Table~\ref{tab7}, when using a hard voting strategy, CoHisNet-tiny achieves 95.06\% ACC and 95.34\% F1, outperforming all competing methods. These results provide complementary evidence that the compact CoHisNet architecture remains competitive on a histopathological dataset with a different tumor type and imaging setting.

\begin{figure}[t]
    \centering
    \includegraphics[width=\columnwidth]{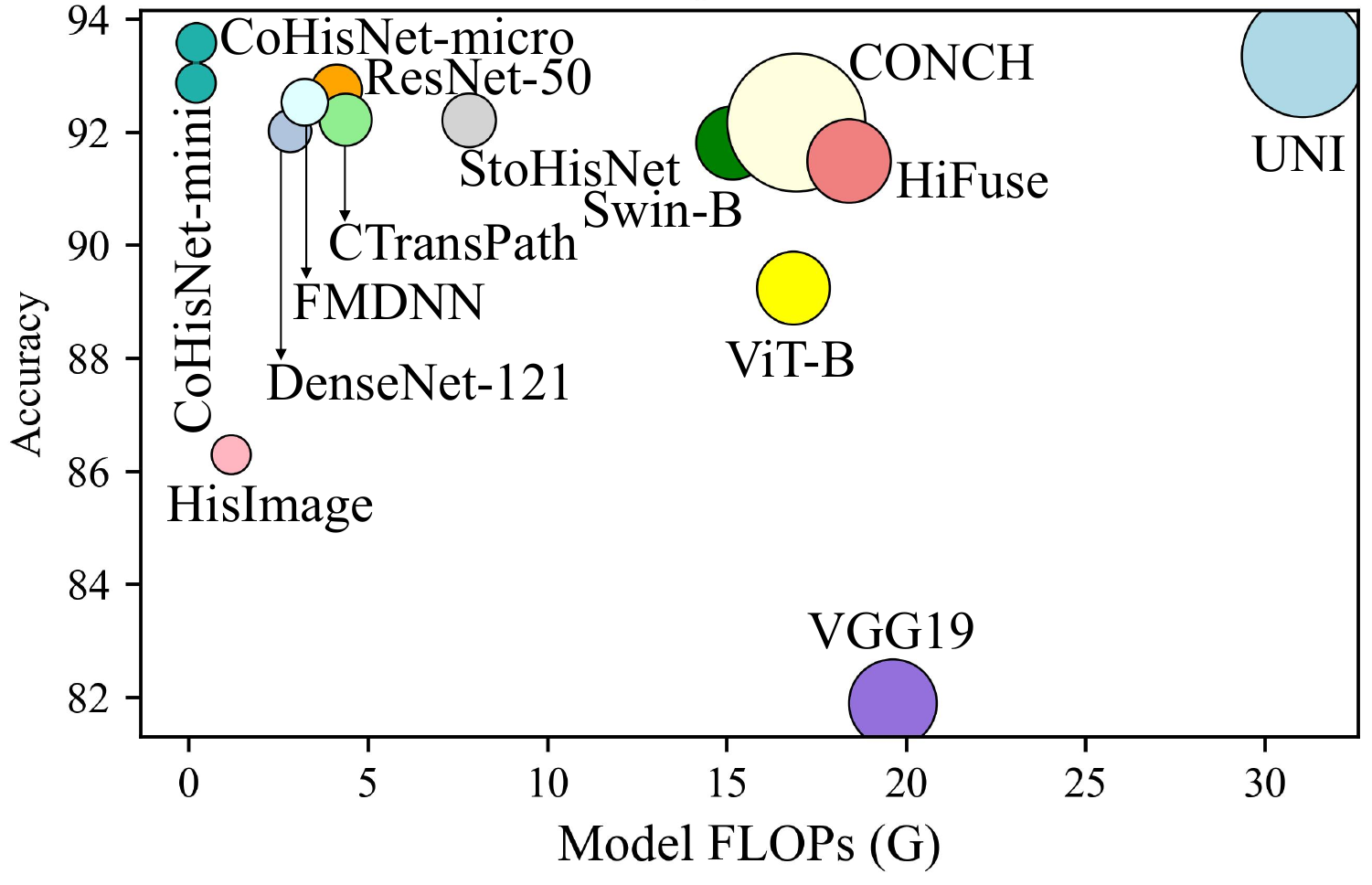} 
    \caption{Relationship between ACC, parameters, and FLOPs of CoHisNet and other advanced models on our private PpNTs dataset. The size of each circle represents the number of parameters for each model.}
    \label{fig:flop}
\end{figure}

\subsubsection{Computational Efficiency}

\begin{table}[t]
    \centering
    \caption{Comparison of parameters and computational complexity between CoHisNet and other models.
}
    \label{tab:pa}
    \begin{tabularx}{\columnwidth}{
        >{\raggedright\arraybackslash}p{3.0cm} 
        >{\centering\arraybackslash}X 
        >{\centering\arraybackslash}X 
    }
    \toprule
    \textbf{Model} & \textbf{Params(M)} & \textbf{FLOPs(G)} \\
    \midrule
    VGGNet~\cite{vgg} & 139.59 & 19.63 \\
    ResNet-50~\cite{resnet} & 23.52 & 4.13 \\
    DenseNet~\cite{densnet} & 6.96 & 2.83 \\
    Vision Transformer~\cite{visiontransformer} & 85.80 & 16.86 \\
    Swin Transformer~\cite{swintransformer} & 86.75 & 15.17 \\

    \midrule
    UNI~\cite{chen2024towards} & 303.35 & 31.06 \\
    UNI2~\cite{chen2024towards} & 681.38 & 185.57\\
    CTransPath~\cite{wang2022transformer} & 27.52 & 4.38 \\
    CONCH~\cite{lu2024visual} & 395.39 & 16.94 \\
    KEEP~\cite{ZHOU2026777} & 414.21 & 61.56 \\

    \midrule
    HisImage~\cite{ding2023enhanced} & 0.53 & 1.19 \\
    FMDNN~\cite{ding2024fmdnn} & 15.21 & 3.24 \\
    StoHisNet~\cite{fu2022stohisnet} & 31.13 & 7.82 \\
    HiFuse~\cite{huo2023hifuse}& 123.27 & 18.41 \\

    \midrule
    \textbf{CoHisNet-mini} & 1.29 & 0.21 \\
    \textbf{CoHisNet-micro} & 1.32 & 0.22 \\
    \textbf{CoHisNet-tiny} & 5.52 & 1.08 \\
    \bottomrule
    \end{tabularx}
\end{table}

CoHisNet achieves a favorable accuracy-complexity trade-off through compact embedding dimensions, reduced stage depth, and KAN-based replacement of conventional MLP sublayers. As illustrated in Fig.~\ref{fig:flop}, CoHisNet-micro achieves the highest internal-test ACC on PpNTs with only 1.32\, M parameters and 0.22\, G FLOPs, while CoHisNet-mini further reduces the computational cost to 0.21\, G FLOPs with only a marginal decrease in accuracy.

Table~\ref{tab:pa} compares the parameter counts and FLOPs of evaluated models. Although HisImage contains fewer parameters than CoHisNet-mini, it requires more FLOPs and yields lower performance on both datasets. These results indicate that CoHisNet is not the smallest model but provides a balance between performance and computational complexity.


\begin{table*}[h]
    \centering
    \scriptsize
    \caption{Ablation studies of CoHisNet on our private PpNTs and public BreakHis datasets.}
    \label{tabam}
    \begin{tabularx}{\textwidth}{
        >{\raggedright\arraybackslash}p{1.35cm}
        >{\centering\arraybackslash}p{0.75cm}
        >{\centering\arraybackslash}p{0.75cm}
        >{\centering\arraybackslash}X
        >{\centering\arraybackslash}X
        >{\centering\arraybackslash}X
        >{\centering\arraybackslash}X
        >{\centering\arraybackslash}X
        >{\centering\arraybackslash}X
        >{\centering\arraybackslash}X
        >{\centering\arraybackslash}X
        >{\centering\arraybackslash}X
        >{\centering\arraybackslash}X
    }
    \toprule
    \multirow{2.5}{*}{\textbf{Model}} 
    & \multirow{2.5}{*}{\textbf{Params}} 
    & \multirow{2.5}{*}{\textbf{FLOPs}} 
    & \multicolumn{5}{c}{\textbf{PpNTs Internal test set}} 
    & \multicolumn{5}{c}{\textbf{BreakHis}}\\
    \cmidrule(lr){4-8} \cmidrule(lr){9-13}
    & & 
    & \textbf{ACC} & \textbf{BACC} & \textbf{Kappa} & \textbf{F1} & \textbf{AUROC} 
    & \textbf{ACC} & \textbf{BACC} & \textbf{Kappa} & \textbf{F1} & \textbf{AUROC}\\
    \midrule
    w/o KAN  & 2.42M & 0.43G & 92.29 & 92.09 & 90.93 & 92.28 & 99.39 & 80.66 & 79.59 & 77.31 & 80.90 & 97.81\\
    w/o CMIF & 1.12M & 0.13G & 92.50 & 92.24 & 90.85 & 92.49 & 99.42 & 79.64 & 81.40 & 77.44 & 80.06 & 97.78\\
    w/o CE   & 1.12M & 0.13G & 92.89 & 92.73 & 90.93 & 92.88 & 99.44 & 82.18 & 82.41 & 75.75 & 82.60 & 98.14\\
    w/o MDWA & 1.32M & 0.22G & \underline{92.98} & \underline{92.77} & \underline{91.91} & \underline{92.96} & \underline{99.47} & \underline{83.46} & \underline{84.72} & \underline{80.75} & \underline{83.89} & \underline{98.45}\\
    \textbf{CoHisNet} & \textbf{1.32M} & \textbf{0.22G} & \textbf{93.58} & \textbf{93.42} & \textbf{92.12} & \textbf{93.57} & \textbf{99.53} & \textbf{87.79} & \textbf{87.56} & \textbf{85.69} & \textbf{88.04} & \textbf{98.70}\\
    \bottomrule
\end{tabularx}
\end{table*}

\subsection{Ablation Experiments}
Our ablation experiments analyzed the contributions of the KAN layer, CMIF module, CE component, and MDWA module across PpNTs and BreakHis. As shown in Table~\ref{tabam}, removing any of these components led to performance degradation, supporting their respective roles in the proposed architecture.

\textbf{Impact of the KAN layer on nonlinear modeling:}
Replacing the KAN layers with conventional MLP layers increased the parameters and FLOPs from 1.32M/0.22G to 2.42M/0.43G, yet reduced performance on both datasets. On PpNTs, ACC decreased from 93.58\% to 92.29\%, while on BreakHis, F1 decreased from 88.04\% to 80.90\%. This suggests that the improvement is not attributable to greater model capacity and that KAN-based nonlinear modeling better supports fine-grained morphological representation.


\textbf{Contribution of the CMIF module to multi-scale fusion:}
Removing CMIF reduced PpNTs BACC from 93.42\% to 92.24\% and Kappa from 92.12\% to 90.85\%. On BreakHis, Kappa decreased from 85.69\% to 77.44\%. This performance drop indicates that cross-scale feature interaction is important for integrating local cellular morphology with broader tissue context. This suggests that CMIF reduces sensitivity to morphology shifts by preserving multi-level information.

\textbf{Role of the CE component in multi-scale enhancement:}
Replacing the CE component with a linear layer decreased PpNTs Kappa from 92.12\% to 90.93\% and F1 from 93.57\% to 92.88\%. On BreakHis, BACC decreased from 87.56\% to 82.41\%. These results suggest that contrast-driven feature enhancement improves the integration of semantic and fine-grained morphological information.

\textbf{Role of the MDWA module in adaptive scale weighting:}
Removing MDWA reduced PpNTs ACC from 93.58\% to 92.98\% and BreakHis Kappa from 85.69\% to 80.75\%. This confirms that learnable scale weighting helps balance features from different stages, although its contribution is smaller than that of the full CMIF design.

\section{Discussion}

Compared with large-scale pathology foundation models~\cite{chen2024towards,wang2022transformer,lu2024visual} and task-specific histopathological classifiers~\cite{ding2023enhanced,ding2024fmdnn,fu2022stohisnet,huo2023hifuse}, CoPath is tailored to the fine-grained and spatially heterogeneous morphology of pNTs. The CMIF module integrates hierarchical representations through adaptive scale weighting and contrast-driven enhancement, allowing cellular-level details to be interpreted together with broader tissue organization. The ablation results indicate that both cross-scale interaction and KAN-based nonlinear transformations contribute to patch-level discrimination, particularly for morphologically similar subtypes. At the WSI level, PathVote further improves upon equal-weight aggregation by assigning greater influence to confident patches containing diagnostically relevant tissue components. These findings suggest that the performance of CoPath arises from the coordinated modeling of local morphology, cross-scale context, and pathology-informed aggregation, rather than from model scale alone.

CoPath is intended as a WSI-level decision-support method and does not replace the reference pathological diagnosis. PathVote incorporates neuropil and Schwannian stromal information as probabilistic priors, whereas complete INPC assessment also considers neuroblastic differentiation, tissue architecture, stromal proportion, and, for selected entities, macroscopic specimen findings. Accordingly, the present retrospective study does not establish clinical benefit; prospective reader studies are required to evaluate its effects on diagnostic accuracy, interobserver agreement, and reporting efficiency.



This study has several limitations. First, although the external validation cohort was scanned with a different device and magnification, it was collected from another branch of the same hospital system; multi-institutional validation is still needed. Second, PathVote relies on pNT-specific histomorphological priors, especially neuropil and Schwannian stroma, which may not directly transfer to other tumor types without redefining the corresponding tissue components. Third, the study focuses on H\&E morphology and does not incorporate molecular markers, genomic alterations, or clinical outcome variables; while WSI-level classification was adopted to ensure label fidelity and interpretability under tumor heterogeneity, patient-level diagnosis remains the clinical goal. Future work will include larger multi-center validation, prospective reader-assistance evaluation, integration with molecular pathology, and development of a pathology diagnostic agent that synthesizes multi-slide WSI predictions with clinical context to achieve comprehensive patient-level reasoning.

\section{Conclusion}
This paper proposed a method named CoPath, which consists of CoHisNet and PathVote, for the pathological-grade diagnosis of NB pathological slices stained with H\&E. The CoHisNet model innovatively integrates multi-scale feature interaction and contrast-driven feature enhancement, effectively addressing the key issues such as insufficient feature integration of existing methods, poor model interpretability, and the easy contamination of multi-scale features by low-density information. Additionally, a heuristic soft voting mechanism was designed, termed PathVote, guided by clinical insights, to achieve an efficient transition from patch-level classification to WSI-level classification. Experiments on private PpNTs datasets, which contain an internal test set and an external validation set, and public BreakHis datasets demonstrate that this method achieved competitive or superior performance on several evaluated metrics. In particular, the lightweight and high-efficiency design of the model offers the possibility of resource-constrained deployment of automated diagnostic medical resources. Future work will focus on multi-center external validation and exploring the integration with molecular diagnostic information to further enhance the clinical utility.


\small
\bibliography{ref}
\end{document}